\documentclass{article}

\usepackage{microtype}
\usepackage{graphicx}
\usepackage{subfigure}
\usepackage{booktabs} 
\usepackage{mathtools}
\usepackage{threeparttable}
\usepackage{pifont}
\newcommand{\cmark}{\ding{51}}
\newcommand{\xmark}{\ding{55}}

\usepackage[colorinlistoftodos,bordercolor=orange,backgroundcolor=orange!20,linecolor=orange,textsize=scriptsize]{todonotes}

\newcommand{\squeeze}{\textstyle}

\usepackage{colortbl}
\definecolor{bgcolor}{rgb}{1, 0.86, 0.69}

\usepackage{hyperref}

\usepackage[accepted]{icml2021}

\icmltitlerunning{Federated Random Reshuffling with Compression and Variance Reduction}
\usepackage{amsmath}
\usepackage{amsfonts}
\usepackage{amsthm}
\usepackage{mathtools}

\newtheorem{proposition}{Proposition}
\newtheorem{definition}{Definition}
\newtheorem{theorem}{Theorem}
\newtheorem{corollary}{Corollary}
\newtheorem{lemma}{Lemma}
\newtheorem{assumption}{Assumption}

\begin{document}

\twocolumn[
\icmltitle{Federated Random Reshuffling with Compression and Variance Reduction}



\icmlsetsymbol{equal}{*}

\begin{icmlauthorlist}
\icmlauthor{Grigory Malinovsky}{to}
\icmlauthor{Peter Richt\'arik}{to}
\end{icmlauthorlist}

\icmlaffiliation{to}{King Abdullah University of Science and Technology KAUST}

\icmlcorrespondingauthor{Grigory Malinovsky}{grigorii.malinovskii@kaust.edu.sa}

\icmlkeywords{Machine Learning, ICML}

\vskip 0.3in
]



\printAffiliationsAndNotice{} 

\begin{abstract}
Random Reshuffling (RR), which is a variant of Stochastic Gradient Descent (SGD) employing sampling {\em without} replacement, is an immensely popular method for training supervised machine learning models via empirical risk minimization. Due to its superior practical performance, it is embedded and often set as default in standard machine learning software. Under the name FedRR, this method was recently shown to be applicable to federated learning \citep{mishchenko2021proximal}, with superior performance when compared to common baselines such as Local SGD. Inspired by this development, we design three new algorithms to improve FedRR further: compressed FedRR and two variance reduced extensions: one for taming the variance coming from shuffling and the other for taming the variance due to compression. The variance reduction mechanism for compression allows us to eliminate dependence on the compression parameter, and applying additional controlled linear perturbations for Random Reshuffling, introduced by~\citet{malinovsky2021random} helps to eliminate variance at the optimum. We provide the first analysis of compressed local methods under standard assumptions without bounded gradient assumptions and for heterogeneous data, overcoming the limitations of the compression operator. We corroborate our theoretical results with experiments on synthetic and real data sets.
\end{abstract}
\section{Introduction}
The primary approach for training supervised machine learning models in the modern machine learning world is Empirical Risk Minimization. While the ultimate goal of supervised learning is to train models that generalize well to unseen data, in practice, only a finite data set is available during training. ERM formulation leads to the following finite-sum optimization problem:
\begin{align}
	\label{finite_sum}
	\min _{x \in \mathbb{R}^{d}} \left[ f(x) = \frac{1}{M} \sum_{m=1}^{M} g_{m}(x)\right],
\end{align}
where each function we also have finite-sum structure
\begin{align*}
	g_m = \frac{1}{n} \sum_{i=1}^{n} f_{m,i}(x).
\end{align*}
Big machine learning models are typically trained in a distributed setting. The training data is distributed across several workers, which compute local updates and then communicate them to the server. We are particularly interested in the Federated Learning setting. Federated Learning~\citep{konevcny2016federated} is a subarea of distributed machine learning, where the number of devices $n$ is enormous. Usually, millions of local devices are heterogeneous to local data and computational and memory resources. Also, users want to keep their privacy, so the algorithm should do training locally. Moreover, communication between workers should be conducted via a trusted aggregation server, which is very expensive.

{\bf Communication as the bottleneck.} 
In literature, we have two strategies to overcome communication issues in federated learning. The first one is communication compression, where our goal is to reduce the number of communicated bits using gradient compression scheme~\citep{mishchenko2019distributed,gorbunov2020unified} and compressed iterates \citep{khaled2019gradient,chraibi2019distributed}. There are many compression techniques such as quantization~\citep{alistarh2017qsgd,bernstein2018signsgd,ramezani2019nuqsgd}, sparsification~\citep{aji2017sparse,lin2017deep,wangni2017gradient,alistarh2018convergence} and other approaches~\citep{shamir2014communication, vogels2019powersgd,wu2018error}. The second strategy to tackle this issue is increasing the number of local steps between the communication rounds. The most popular algorithm --- FedAvg~\citep{mcmahan2017communication}--- is based on this idea. Many papers provide theoretical justifications for special cases of FedAvg such as local GD~\citep{khaled2019first} and local SGD~\citep{khaled2020tighter,gorbunov2020unified,stich2018local, lin2018don}. The natural union of communication compression and local computations is presented in~\citet{Basu2020QsparseLocalSGDDS,haddadpour2021federated}. However, the theory provided in these papers is limited due to unrealistic assumptions. 

{\bf Sampling without replacement.} 
Stochastic first-order algorithms, in particular, have attracted much attention in the machine learning world. Of these, stochastic gradient descent (SGD) is perhaps the best known and the most basic. SGD has a long history~\citep{robbins1951stochastic} and is therefore well-studied and well-understood~\citep{gower2019sgd}. However, methods based on data permutations~\citep{bottou2009curiously}, when data points are shuffled randomly and processed in order, without replacement, show better performance than SGD~\cite{recht2013parallel}. Also, this method has software implementation advantages since these methods are friendly to cache locality~\citep{Bengio2012}. The most popular model in this class is Random Reshuffling~\citep{recht2012toward}. Shuffle-Once~\citep{safran2020good} uses a similar approach, but shuffling occurs only once, at the very beginning, before the training begins. Random Reshuffling and Shuffle-Once have a long history, and many theoretical works try to show the advantages of Random Reshuffling~\citep{gurbuzbalaban2019random,haochen2018random,Nagaraj2019} and Shuffle Once~\citep{Rajput2020}.

{\bf Federated Random Reshuffling.} Recent advances of \citet{mishchenko2020random} and providing extension for Federated Learning \citep{mishchenko2021proximal} allow us to consider this technique as a particular variant of FedAvg with a fixed number of local computations and sampling without replacement. 

{\bf Variance Reduction.} Compression operators help to reduce the number of transmitted bits. However, at the same time, it starts to be a source of variance, which increases the neighborhood of the optimal solution. This variance can sufficiently slow down the algorithm. In order to overcome this challenge, we need to use a variance reduction mechanism. The idea of this approach is based on shifted compression operator, and firstly it was proposed for compressed gradients~\citep{mishchenko2019distributed}. For compressed iterates, the variance reduction mechanism was proposed in \citet{chraibi2019distributed}. Moreover, stochastic first-order methods become a source of variance due to their random nature. Hopefully, another variance reduction mechanism can help with this type of variance. There are many variance-reduced methods which use sampling with replacement such as SVRG~\citep{johnson2013accelerating}, L-SVRG~\citep{kovalev2020don},SAGA~\citep{defazio2014saga},SAG~\citep{roux2012stochastic},Finito\citep{pmlr-v32-defazio14} etc. For permutation-based algorithms, we have only a few variance-reduced methods~\citep{Ying2019,park2020linear,mokhtari2018surpassing}. A recent paper of \citet{malinovsky2021random} introduced linear perturbation reformulation that allows getting better rates for variance reduced Random Reshuffling. 

\section{Contributions}
This section outlines our work's key contributions and offers explanations and clarifications regarding some of the development. 

{\bf Compressed FedRR.} We propose the first method, which combines three ideas: compression, local steps, and sampling without replacement. This method is compressed federated random reshuffling. The basic approach is to apply the compression operator to the iterates after each epoch and then aggregate compressed updates. Applying compression to the iterates can significantly worsen convergence properties. We prove the following rate:

\begin{align*}
	\squeeze
	&\mathbb{E}\|x_T-x_*\|^2\leq (1-\gamma\mu)^{\frac{nT}{2}}\mathbb{E}\|x_0-x_*\|^2\\
	&+\frac{2\omega}{M}\frac{1}{\gamma\mu}\frac{1}{M}\sum_{m=1}^{M}\|x^n_{*,m}\|^2\\
	&+\frac{2}{\mu}\left(1+\frac{2\omega}{M}\right)\gamma^2L\frac{1}{M}\sum_{m=1}^{M}\left(\|\nabla F_m(x_*)\|^2+\frac{n}{4}\sigma_{*,m}^2\right)
\end{align*}
As we can see, we have a part with linear rate:  $(1-\gamma\mu)^{\frac{nT}{2}}\mathbb{E}\|x_0-x_*\|^2$. Also there are three sources of variance in the optimum. The first one is $\frac{2\omega}{M}\frac{1}{\gamma\mu}\frac{1}{M}\sum_{m=1}^{M}\|x^n_{*,m}\|^2$ and it caused by compression. It is equal to zero only if $\omega = 0$. This term cannot be elimated by decreasing step-sizes strategies. The second term is $\frac{2}{\mu}\left(1+\frac{2\omega}{M}\right)\gamma^2L\frac{1}{M}\sum_{m=1}^{M}\frac{n}{4}\sigma_{*,m}^2$. This source of variance is caused by stochasticity of Random Reshuffling method. This variance can be decreased by decreasing step-sizes. The third term is $\frac{2}{\mu}\left(1+\frac{2\omega}{M}\right)\gamma^2L\frac{1}{M}\sum_{m=1}^{M}\|\nabla F_m(x_*)\|^2$. This source of variance is caused by heterogeneity of data. In other words, if we have the same optimum for all functions $F_m(x)$, we can get rid of this term. In heterogenious regime we need to use additional mechanism for controlling client drift such as SCAFFOLD~\cite{Karimireddy2019}.

{\bf Variance-reduced Compressed FedRR.} In the previous section, it was shown that using compressed iterates causes additional variance, which cannot be vanished by decreasing step-sizes strategies. Moreover, it forces us to have an additional assumption on the compression operator. We need to have very small compression parameter $\omega \leq \frac{M\gamma\mu\varepsilon}{\frac{2}{M}\|x^n_{*,m}\|^2}$. In order to fix it, we propose Variance-reduced compressed FedRR (FedCRR-VR) that utilizes shifted compressed updates with learning shifts. The similar mechanism is used in \citet{mishchenko2019distributed} and \citet{gorbunov2020unified}.

{\bf Double Variance-reduced Compressed FedRR.}
We propose a modification of Variance-reduced Compressed Federated Random Reshuffling, which allows eliminating variance caused by stochasticity. We armed Variance-reduced Compressed FedRR with the linear permutation approach proposed by \citet{malinovsky2021random}. Now we get both variance reduction mechanisms in one algorithm.

	{\footnotesize
	\begin{table*}
		\caption{Comparison of the main features of our algorithms and results with other methods with compression.} \label{tbl:main}
		\centering
		\tiny
		\begin{threeparttable}
			
			\renewcommand{\arraystretch}{2}
			\begin{tabular}{|p{18mm}|p{12mm}|p{11mm}|p{11mm}|p{4mm}|p{12mm}|p{42mm}|}
				
				\hline
				Feature &    Variance reduction for compression  & Variance reduction for stochasticity  & Variance reduction for client-drift & RR\tnote{\color{red} (1)} &  No additional assumption &  Communication complexity\tnote{\color{red}(2)} \\
				\hline
				\hline
				QSPARSE \citep{Basu2020QsparseLocalSGDDS} & \xmark & \cmark&\cmark& 
				\xmark&\xmark\tnote{\color{red} (3)}&$\mathcal{O}\left(\frac{\kappa(\omega+1)}{\sqrt{\varepsilon}}\right)$\\
				\hline
				FedCOMGATE \citep{haddadpour2021federated}& \cmark & \cmark&\cmark&\xmark&\xmark \tnote{\color{red} (4)}&$\tilde{\mathcal{O}}\left(\kappa(\frac{\omega}{M}+1)\right)$\\
				\hline 
				DIANA\citep{mishchenko2019distributed}&\cmark&\xmark&\xmark&\xmark&\cmark&$O\left(\kappa \max \left\{V^{0}, \frac{\left(1-\alpha_{p}\right) \sigma^{2}}{\mu L n \alpha_{p}}\right\} \frac{1}{\varepsilon}\right)$\\
				\hline
				FedCRR& \xmark & \xmark&\xmark&\cmark&\xmark\tnote{\color{red} (5)} &{\tiny$\tilde{\mathcal{O}}\left( \left( \kappa +\frac{\sqrt{\kappa}(\sigma_{cl}+\sigma_{st})}{\mu\sqrt{\varepsilon}}\right) \right)$}\tnote{\color{red} (6)} \\
				\hline
				FedCRR-VR& \cmark & \xmark&\xmark&\cmark&\cmark&$\tilde{\mathcal{O}}\left(\frac{(\omega+1)\left(1-\frac{1}{\kappa}\right)^n}{\left(1-\left(1-\frac{1}{\kappa}\right)^n\right)^2}+\frac{\sqrt{\kappa}(\sigma_{cl}+\sigma_{st})}{\mu\sqrt{\varepsilon}}\right)$  \\
				\hline
				FedCRR-VR-2& \cmark & \cmark&\xmark&\cmark&\xmark\tnote{\color{red} (7)}& $\tilde{\mathcal{O}}\left(\frac{(\omega+1)\left(1-\frac{1}{\kappa\sqrt{\kappa n}}\right)^{\frac{n}{2}}}{\left(1-\left(1-\frac{1}{\kappa\sqrt{\kappa n}}\right)^{\frac{n}{2}}\right)^2}+\frac{\sqrt{\kappa}(\sigma_{cl})}{\mu\sqrt{\varepsilon}}\right)$ \\

				\hline
				
			\end{tabular}
			\begin{tablenotes}
				{\scriptsize
					\item {\color{red}(1)} Sampling without replacement (Random Reshuffling).
					\item {\color{red}(2)} $\tilde{\mathcal{O}}$ notation ignores logarithmic factors.
					\item {\color{red}(3)} Bounded second moment: $\mathbb{E}\left[\left\|\nabla f_{i}\left(x\right)\right\|_{2}^{2}\right] \leq G_1^{2} $. Contractive compressor assumption: $\mathbb{E}\left[\left\|x-\mathcal{C}(x)\right\|_{2}^{2}\right] \leq(1-\gamma)\|x\|_{2}^{2}$ for $\gamma \in(0,1]$.
					\item {\color{red}(4)}  Assumption that for all  $x_1 , \ldots , x_n \in \mathbb{R}^d$ the compressor $\mathcal{C}$ satisfies $\mathbb{E}\left[\left\|\frac{1}{n} \sum_{i=1}^{n} \mathcal{C}\left(x_{j}\right)\right\|^{2}-\left\|\mathcal{C}\left(\frac{1}{n} \sum_{i=1}^{n} x_{j}\right)\right\|^{2}\right] \leq G_2$. Classical compression operators like RandK and $l_2$-quantization on $\mathbb{R}^d$ does not satisfy this condition. As counterexample we can set $n=2$ and $x_2 = -x_1 = t\cdot (1,1,\ldots,1)^{\top}$ with arbitrary large $t>0$.
					\item {\color{red}(5)} Small compression operator: $\omega \leq \frac{M\gamma\mu\varepsilon}{\frac{2}{M}\|x^n_{*,m}\|^2}$.
					\item{\color{red}(6)}  Client drift: $\sigma_{cl} = \frac{1}{M}\sum_{m=1}^{M}\|F_m(x_*)\|$, sum of local variances: $\sum_{1}^{M}\frac{\sqrt{n}}{2}\sigma_{*,m} $
					\item {\color{red}(7)} Big data regime: $n>\log \left(\frac{1}{1-\delta^{2}}\right)\left(\log \left(\frac{1}{1-\gamma \mu}\right)\right)^{-1}$, where $0<\delta<1$.
				}
			\end{tablenotes}
		\end{threeparttable}
	\end{table*}
}
\section{Preliminaries}
\subsection{$L$-smooth and $\mu$-strongly convex  functions}
Before introducing our convergence results, let us first formulate all concepts that we use throughout the paper. Firstly, we consider a class of $mu$-strongly convex and $L$-smooth functions. 
\begin{definition}
	A differentiable function $f$ is $\mu$-strongly convex if
	$$f(y) \geq f(x)+\nabla f(x)^{T}(y-x)+\frac{\mu}{2}\|y-x\|^{2}$$
	for $\mu>0$ and all $x,y$.
\end{definition}

\begin{definition}
	A differentiable function $f$ is $L$-smooth if
	$$f(y) \leq f(x)+\nabla f(x)^{T}(y-x)+\frac{L}{2}\|y-x\|^{2}$$
	for some $L>0$ and all $x,y$.
\end{definition}
There is the first assumption that we use in all theorems.
\begin{assumption}
	\label{assm:f}
	Each $f_{m,i}$ is $\mu$-strongly convex and $L$-smooth. 
\end{assumption}
We also need to define Bregman divergence, which is often used in the analysis.
\begin{definition}
	The Bregman divergence with respect to $f$ is the mapping $D_f :\mathbb{R}^d \times \mathbb{R}^d \rightarrow \mathbb{R}$ defined as follows:
	$$D_{f}(x, y)\stackrel{\text { def }}{=}f(x)-f(y)-\langle\nabla f(y), x-y\rangle .$$
\end{definition}

\subsection{Compression operator}
In order to overcome communication issues, we apply a compression operator to the iterates. Now we are going to extend Federated Random Reshuffling using compression. Let us define the concept of compressors. 
\begin{definition}
	We say that a randomized map $\mathcal{C}:\mathbb{R}^d\to \mathbb{R}^d$ is in class $ \mathbb{B}^{d}(\omega)$ if  there exists a constant $\omega \geq 0$ such that the following relations hold for all $x\in \mathbb{R}^d$:
	\begin{equation*}\label{chap-compr:def-compressor}
		\mathbb{E}\left[ \mathcal{C}(x)\right]  = x ,\quad
		\mathbb{E}  \left[\|\mathcal{C}(x) \|^2 \right] \leq (\omega + 1)\|x\|^2.
	\end{equation*} 
\end{definition}
\begin{assumption}
	\label{assm:compre}
	All compression operators are in class $ \mathbb{B}^{d}(\omega)$.
\end{assumption}
This class of compressor operators is classical in literature~\citep{mishchenko2019distributed,horvath2019natural,Basu2020QsparseLocalSGDDS}.
\subsection{Random Reshuffling, Shuffle Once}
In order to conduct analysis for sampling without replacement we need to establish specific notions. We sample a random permutation $\left\{\pi_{0}, \pi_{1}, \ldots, \pi_{n-1}\right\}$ of the set $\{1,2, \ldots, n\}$, and proceed with $n$ iterates of the form
$x_{t,m}^{i+1}=x_{t,m}^{i}-\gamma \nabla f_{m,\pi_{i}}\left(x_{t,m}^{i}\right)$ at each machine locally. We also consider option when we have only one random permutation, at the very beginning, and then algorithm uses this permutation during the whole process. 

For a constant stepsize and a fixed permutation, we define intermediate limit point:
\begin{align*}
	\label{x_*_n}
	x_{*}^{i} \stackrel{\text { def }}{=} x_{*}-\gamma \sum_{j=0}^{i-1} \nabla f_{\pi_{j}}\left(x_{*}\right), \quad i=1, \ldots, n-1.
\end{align*}
To measure the closeness between $x_*$ and $x_{*}^n$ we use definition from~\citet{mishchenko2021proximal} of Shuffling radius.
\begin{definition}
	For given a stepsize $\gamma>0$ and a random permutation $\pi$ of $\{1,2, \ldots, n\}$ shuffling radius is defined by 
	$$\sigma_{\mathrm{rad}}^{2}\stackrel{\text { def }}{=} \max _{i=1, \ldots, n-1}\left[\frac{1}{\gamma^{2}} \mathbb{E}_{\pi}\left[D_{f_{\pi_{i}}}\left(x_{*}^{i}, x_{*}\right)\right]\right].$$
\end{definition}
We also need to define the most popular parameter for method's stochastisity.
\begin{definition}
	Variance at the optimum:
	\begin{align*}
		\sigma_{*}^{2} \stackrel{\text { def }}{=} \frac{1}{n} \sum_{i=1}^{n}\left\|\nabla f_{i}\left(x_{*}\right)-\nabla f\left(x_{*}\right)\right\|^{2}.
	\end{align*}
\end{definition}
The shuffling radius for permutation-based algorithms is natural, and it is more convenient to work with this concept. However, we need to have an upper bound in terms of $\sigma^2_*$ to compare different methods. To get an upper bound for shuffling radius, we need to use a lemma in~\citet{mishchenko2020random} that bounds variance of sampling without replacement. 
\begin{theorem}
	\label{thm:rad}
	For any stepsize $\gamma>0$ and any random permutation $\pi$ of $\{1,2\ldots,n\}$ we have 
	$$\sigma_{\mathrm{rad}}^{2} \leq \frac{L_{\max }}{2} n\left(n\left\|\nabla f\left(x_{*}\right)\right\|^{2}+\frac{1}{2} \sigma_{*}^{2}\right).$$
\end{theorem}
In case when we have only one node we obtain that $\left\|\nabla f\left(x_{*}\right)\right\|^{2} = 0$. However, in multuple node case we will need this term.
\subsection{Lifted problem reformulation}
Let us consider a bigger product space by introducing dummy variables and the constraint $x_1 = x_2 = \ldots = x_M$. We need to define regularizer for this reformulation:
\begin{align*}
	\psi\left(x_{1}, \ldots, x_{M}\right)=\left\{\begin{array}{ll}
		0, & x_{1}=\cdots=x_{M} \\
		+\infty, & \text { otherwise .}
	\end{array}\right.
\end{align*}
Using this regularizer, we can establish the reformulated problem:
\begin{align*}
	\min _{x_{1}, \ldots, x_{M} \in \mathbb{R}^{d}} \frac{1}{nM} &\sum_{m=1}^{M} F_{m}\left(x_{m}\right)+\psi\left(x_{1}, \ldots, x_{M}\right)\\
	F_{m}(x)=&\sum_{j=1}^{n} f_{m j}(x).
\end{align*}
We need to have an upper bound for the shuffling radius for this reformulated problem. First, we need to define the variance of the method's stochasticity in distributed case.
\begin{definition}
	The variance of local gradients:
	\begin{align*}
		\sigma_{m, *}^{2} \stackrel{\text { def }}{=} \frac{1}{n} \sum_{j=1}^{n}\left\|\nabla f_{m j}\left(x_{*}\right)-\frac{1}{n} \nabla F_{m}\left(x_{*}\right)\right\|^{2}.
	\end{align*}
\end{definition}
Now we need to use a lemma from~\citep{mishchenko2021proximal} to bound shuffled radius for the reformulated problem:
\begin{lemma}
	\label{lemma:rad}
	The shuffling radius $\sigma^2_{rad}$ of lifted problem is upper bounded by 
	\begin{align*}
		\sigma_{\mathrm{rad}}^{2} \leq L \sum_{m=1}^{M}\left(\left\|\nabla F_{m}\left(x_{*}\right)\right\|^{2}+\frac{n}{4} \sigma_{m, *}^{2}\right).
	\end{align*}
\end{lemma}
We can see that there are two parts of variance. First one depends on the sum of local variances $\sum_{m=1}^{M} \sigma_{m, *}^{2}$. The second part depends on sum of local gradient norms $\sum_{m=1}^{M}\left\|\nabla F_{m}\left(x_{*}\right)\right\|^{2}$. Both of these terms appear in analysis of local SGD~\cite{khaled2020tighter}.

\section{Compressed Federated Random Reshuffling}
In this section, we propose a direct application of compressed iterates to federated Random Reshuflling. In this procedure server distributes the current point to workers, then each worker computes the full epoch according to its sampled permutation locally. After that, the final iterate $x^n_{t,m}$ is compressed and transmitted to the server, where all updates are aggregated by taking the average. 
\begin{theorem}
	\label{thm:CRR}
	Suppose that Assumption~\ref{assm:compre} and Assumption~\ref{assm:f} hold. Additionally assume that compression parameter is sufficiently small: $\omega\leq \frac{M}{2}\frac{1-\left(1-\gamma\mu\right)^{\frac{n}{2}}}{\left(1-\gamma\mu\right)^{\frac{n}{2}}}$. If the stepsize satisfies $\gamma \leq \frac{1}{L}$, the iterates generated by FedCRR or FedCSO (Algorithm~\ref{alg:fed_rr}) satisfy
	\begin{align*}
		\squeeze 
		&\mathbb{E} \left[\|x_{t+1} - x_*\|^2\right] \leq (1-\gamma\mu)^{\frac{nT}{2}}\|x_{0} - x_*\|^2\\
		&+\frac{2}{\mu}\gamma^2L_{\max}\frac{1}{M}\sum_{m=1}^{M}\left(\|F_m(x_*)\|^2+\frac{n}{4}\sigma_{*,m}^2\right)\\
		&+\frac{2\omega}{M}\frac{1}{\gamma\mu}\frac{1}{M}\sum_{m=1}^{M}\|x^n_{*,m}\|^2.
	\end{align*}
\end{theorem}
\begin{algorithm}[t]
	\caption{Federated Compressed Random Reshuffling (FedCRR) and Shuffle-Once (FedCSO)}
	\label{alg:fed_rr}
	\begin{algorithmic}[1]
		\STATE \textbf{Parameters:} Stepsize $\gamma > 0$, initial vector $x_0 = x_0^0 \in \mathbb{R}^d$, number of epochs $T$
		\STATE For each $m$, sample permutation $\pi_{0, m}, \pi_{1, m}, \ldots, \pi_{n-1, m}$ of $\{ 1, 2, \ldots, n \}$ (Only FedCSO)
		\FOR{epochs $t=0,1,\dotsc,T-1$}
		\FOR{$m=1,\dotsc, M$ locally in parallel}
		\STATE $x_{t, m}^0=x_t$
		\STATE Sample permutation $\pi_{0, m}, \pi_{1, m}, \ldots, \pi_{n-1, m}$ of $\{ 1, 2, \ldots, n \}$ (Only FedCRR)
		\FOR{$i=0, 1, \ldots, n-1$}
		\STATE $x_{t, m}^{i+1} = x_{t, m}^{i} - \gamma \nabla f_{\pi_{i, m}} (x_{t, m}^i)$
		\ENDFOR
		\STATE $q_{t, m} = \mathcal{C}(x_{t,m}^{n})$
		\ENDFOR
		\STATE $x_{t+1}=\frac{1}{M}\sum_{m=1}^M q_{t, m}$;  
		\ENDFOR
	\end{algorithmic}
\end{algorithm}
\begin{algorithm}[t]
	\caption{Variance Reduced Federated Compressed Random Reshuffling (FedCRR-VR) and Shuffle-Once (FedCSO-VR)}
	\label{alg:vr_fed_rr}
	\begin{algorithmic}[1]
		\STATE \textbf{Parameters:} Stepsize $\gamma > 0$, initial vector $x_0 = x_0^0 \in \mathbb{R}^d$, number of epochs $T$
		\STATE For each $m$, sample permutation $\pi_{0, m}, \pi_{1, m}, \ldots, \pi_{n-1, m}$ of $\{ 1, 2, \ldots, n \}$ (Only FedCSO-VR)
		\FOR{epochs $t=0,1,\dotsc,T-1$}
		\FOR{$m=1,\dotsc, M$ locally in parallel}
		\STATE $x_{t, m}^0=x_t$
		\STATE Sample permutation $\pi_{0, m}, \pi_{1, m}, \ldots, \pi_{n-1, m}$ of $\{ 1, 2, \ldots, n \}$ (Only FedCRR-VR)
		\FOR{$i=0, 1, \ldots, n-1$}
		\STATE $x_{t, m}^{i+1} = x_{t, m}^{i} - \gamma \nabla f_{\pi_{i, m}} (x_{t, m}^i)$
		\ENDFOR
		\STATE $q_{t, m} = \mathcal{C}(x_{t,m}^{n} - h_{t,m})$
		\STATE $h_{t+1,m} = h_{t,m} +\alpha q_{t, m} $
		\ENDFOR
		\STATE $x_{t+1}=(1-\eta)x_t+\eta\frac{1}{M}\sum_{m=1}^M\left(q_{t, m} +h_{t,m}\right) $;  
		\ENDFOR
	\end{algorithmic}
\end{algorithm}

We can see that the last term makes the largest contribution to the size of the neighborhood, and decreasing stepsizes cannot help. Now we establish communication complexity.
\begin{corollary}
	
	Let the assumptions in the Theorem~\ref{thm:CRR} hold. Also assume that
	$\omega \leq \frac{M\gamma\mu\varepsilon}{\frac{2}{M}\|x^n_{*,m}\|^2}$.
	Then the communication complexity of Algorithm~\ref{alg:fed_rr} is 
	$$
	\squeeze
	T = \tilde{\mathcal{O}}\left( \left( \kappa +\frac{\sqrt{\kappa}}{\mu\sqrt{\varepsilon}}\Delta\right)\log \left(\frac{1}{\varepsilon}\right) \right),$$
	where $\Delta = \frac{1}{M}\sum_{m=1}^{M}\left(\|\nabla F_m(x_*)\|+\sqrt{n}\sigma_{*,m}\right)$
\end{corollary}
\begin{algorithm}[t]
	\caption{Double Variance Reduced Federated Compressed Random Reshuffling (FedCRR-VR-2) and Shuffle-Once (FedCSO-VR-2)}
	\label{alg:dvr_fed_rr}
	\begin{algorithmic}[1]
		\STATE \textbf{Parameters:} Stepsize $\gamma > 0$, initial vector $x_0 = x_0^0 \in \mathbb{R}^d$, number of epochs $T$
		\STATE For each $m$, sample permutation $\pi_{0, m}, \pi_{1, m}, \ldots, \pi_{n-1, m}$ of $\{ 1, 2, \ldots, n \}$ (Only VR-FedCSO)
		\FOR{epochs $t=0,1,\dotsc,T-1$}
		\FOR{$m=1,\dotsc, M$ locally in parallel}
		\STATE $x_{t, m}^0=x_t$
		\STATE $y_t = x_t$
		\STATE Sample permutation $\pi_{0, m}, \pi_{1, m}, \ldots, \pi_{n-1, m}$ of $\{ 1, 2, \ldots, n \}$ (Only D-VR-FedCRR)
		\FOR{$i=0, 1, \ldots, n-1$}
		\STATE $g\left(x_{t,m}^{i}, y_{t}\right)=\nabla f_{\pi_{i},m}\left(x_{t,m}^{i}\right)-\nabla f_{\pi_{i},m}\left(y_{t}\right)+\frac{1}{n}\nabla F_m \left(y_{t}\right)$		
		\STATE $	x_{t,m}^{i+1}=x_{t,m}^{i}-\gamma g\left(x_{t,m}^{i}, y_{t}\right)$
		\ENDFOR
		\STATE $q_{t, m} = \mathcal{C}(x_{t,m}^{n} - h_{t,m})$
		\STATE $h_{t+1,m} = h_{t,m} +\alpha q_{t, m} $
		\ENDFOR
		\STATE $x_{t+1}=(1-\eta)x_t+\eta\frac{1}{M}\sum_{m=1}^M\left(q_{t, m} +h_{t,m}\right) $;  
		\ENDFOR
	\end{algorithmic}
\end{algorithm} 
\section{Variance Reduced Compressed Federated Random Reshuffling}
In this section, we introduce a variance reduction mechanism for compression in order to upgrade Algorithm~\ref{alg:fed_rr}. The main part of the algorithm remains the same. However, after each epoch, we apply the compression operator to the difference between local iterates and learning shifts. After that, at each node, we compute updates of learning shifts. To control the learning process of shifts, we use additional parameter $\alpha$. To get convergence we need to satisfy $\alpha\leq\frac{1}{\omega+1}$. After that server aggregates updates by using a convex combination of previous iterate and average of updates. To control this convex combination, we use additional parameter $\eta$. The next theorem shows that this mechanism helps to get rid of compression variance and the additional assumptions. To get the convergence rate, we introduce the Lyapunov function.

\begin{theorem}
	\label{thm:VR-CRR}
	Suppose that Assumption~\ref{assm:f} and Assumption~\ref{assm:compre} hold.  Then provided the stepsize satisfies $\gamma \leq \frac{1}{L}$, $\alpha\leq\frac{1}{\omega+1}$ and $\eta\leq \min \left( 1,\frac{\left(1-\left(1-\gamma\mu\right)^n\right)M}{12\omega\left(1-\gamma\mu\right)^n}\right)$ the iterates generated by FedCRR-VR or FedCSO-VR (Algorithm~\ref{alg:vr_fed_rr}) satisfy
	\begin{align*}
		\squeeze
		\mathbb{E}\Psi_T &\leq \left(1-\frac{\min \left(\alpha,\eta(1-(1-\gamma\mu)^n)\right)}{2}\right)^T\Psi_0\\
		&+\frac{2\left(\alpha+\eta+\frac{2\eta^2\omega}{M}\right)\gamma^{3} L_{\max }}{M \left(\alpha,\eta(1-(1-\gamma\mu)^n)\right)} \sum_{m=1}^{M}\delta_m,
	\end{align*}
	where Lyapunov function is defined as $\Psi_t = \|x_t-x_*\|^2+\frac{4 \eta^{2} \omega}{\alpha M}\frac{1}{M}\sum_{m=1}^{M}\left\| h_{t,m} - x^n_{*,m}  \right\|^2$ and $\delta_m = \left(\left\|\nabla F_{m}\left(x_{*}\right)\right\|^{2}+\frac{n}{4} \sigma_{m, *}^{2}\right).$
\end{theorem}
Now there is no compression variance term anymore. Next corollary demonstrates communication complexity. 
\begin{corollary}
	Let the assumptions in the Theorem~\ref{thm:VR-CRR} hold. Then the communication complexity of Algorithm~\ref{alg:vr_fed_rr} is 
	$$
	\squeeze
	T = \mathcal{O}\left(\left(\frac{(\omega+1)\left(1-\frac{1}{\kappa}\right)^n}{\left(1-\left(1-\frac{1}{\kappa}\right)^n\right)^2}+\frac{\sqrt{\kappa}}{\mu\sqrt{\varepsilon}}\Delta\right) \log \left(\frac{1}{\varepsilon}\right) \right),$$
	where $\Delta = \frac{1}{M}\sum_{m=1}^{M}\left(\|\nabla F_m(x_*)\|+\sqrt{n}\sigma_{*,m}\right).$
\end{corollary}
We have the same second term, which depends on the sum of local gradients and local variances. Also, the linear rate is slightly worse since we can use any compression operator, and we also need to learn shifts. However, the main advantage of this method is the possibility of using any compression parameter $\omega$.

\section{Double Variance Reduced Compressed Federated Random Reshuffling}
This section proposes another variance reduction mechanism to eliminate local variances caused by the method's stochasticity. To achieve this goal, we need to use inner product reformulation introduced by~\citet{malinovsky2021random}. We can get an equivalent form of the local function. Let $a_{i}, \ldots, a_{n} \in \mathbb{R}^{d}$ are vectors that sum to zero $\sum_{i=1}^{n} a_{i}=0$:
\label{perty}
\begin{align}
	F_m(x) = \sum_{i=1}^{n} \left( f_{i,m} + \left\langle a_{i,m},x \right\rangle  \right)= \sum_{i=1}^{n} \tilde{f}_{i,m}.
\end{align}
Let us consider the following gradient estimate: 
$$g\left(x_{t}^{i}, y_{t}\right)=\nabla f_{\pi_{i},m}\left(x_{t}^{i}\right)-\nabla f_{\pi_{i},m}\left(y_{t}\right)+\frac{1}{n}\nabla F_m\left(y_{t}\right) .$$
Obviously, the sum of these vectors is equal to zero: 
$$
\sum_{i=1}^{n} a_{i,m}=-\sum_{i=1}^{n} \nabla f_{\pi_{i},m}\left(y_{t}\right)+\frac{1}{n}\sum_{i=1}^{n} \nabla F_m\left(y_{t}\right)=0 .$$
Now we are ready to formulate the theorem of convergence guarantees.
\begin{theorem}
	\label{thm:double}
	Suppose that Assumption~\ref{assm:compre} and Assumption~\ref{assm:f} hold.  Then provided the stepsize satisfies $\gamma \leq \frac{1}{8L}\sqrt{\frac{\mu}{nL}}$, $\alpha\leq\frac{1}{\omega+1}$, $\eta\leq \min \left( 1,\frac{\left(1-\left(1-\gamma\mu\right)^{\frac{n}{2}}\right)M}{12\omega\left(1-\gamma\mu\right)^{\frac{n}{2}}}\right)$ and $\frac{1}{8} \leq(1-\gamma \mu)^{\frac{n}{2}}\left(1-(1-\gamma \mu)^{\frac{n}{2}}\right),$ the iterates generated by FedCRR-VR-2 or FedCSO-VR-2 (Algorithm~\ref{alg:dvr_fed_rr}) satisfy
	\begin{align*}
		\squeeze
		\mathbb{E}\Psi_T &\leq \left(1-\frac{1}{2}\min \left(\alpha, \eta (1-(1-\gamma\mu)^{\frac{n}{2}})\right)\right)^T\Psi_0\\
		&+\frac{2\left(\alpha+\eta+\frac{2\eta^2\omega}{M}\right)\gamma^3 L\sum_{m=1}^{M}\left(\|\nabla F_m(x_*)\|^2\right)}{M \min \left(\alpha, \eta (1-(1-\gamma\mu)^{\frac{n}{2}})\right)},
	\end{align*}
	where Lyapunov function is defined as $\Psi_t = \|x_t-x_*\|^2+\frac{4 \eta^{2} \omega}{\alpha M}\frac{1}{M}\sum_{m=1}^{M}\left\| h_{t,m} - x^n_{*,m}  \right\|^2.$
\end{theorem}
We need to use smaller stepsize since we applied variance reduction mechanism. However, we managed to vanish sum of local variances. The next theorem shows the communication complexity of Algorithm~\ref{alg:dvr_fed_rr}.
\begin{corollary}
	Let the assumptions in the Theorem~\ref{thm:VR-CRR} hold. Then the communication complexity of Algorithm~\ref{alg:dvr_fed_rr} is 
	$$\squeeze
	T = \mathcal{O}\left(\left(\frac{(\omega+1)\left(1-\frac{1}{\kappa\sqrt{\kappa n}}\right)^{\frac{n}{2}}}{\left(1-\left(1-\frac{1}{\kappa\sqrt{\kappa n}}\right)^{\frac{n}{2}}\right)^2}+\frac{\sqrt{\kappa}}{\mu\sqrt{\varepsilon}}\Delta^{\prime}\right)\log \left(\frac{1}{\varepsilon}\right)\right),$$
	where $\Delta^{\prime} = \frac{1}{M}\sum_{m=1}^{M}\left(\|\nabla F_m(x_*)\|\right).$
\end{corollary}

\section{Experiments}
{\bf Model}.
In our experiments we solve the regularized ridge regression problem, which has the form~\ref{finite_sum} with 
$\squeeze f_{im}(x)=\frac{1}{2}\left\|A^m_{i,:} x-y^m_{i}\right\|^{2}+\frac{\lambda}{2}\|x\|^{2},$
where $A^m \in \mathbb{R}^{n \times d}, y^m \in \mathbb{R}^{n}$ and $\lambda>0$ is regularization parameter. Consider concatenated matrix $A \in\mathbb{R}^{mn\times d}$. This problem satisfies Assumption~\ref{assm:f} for $L=\max _{i}\left\|A_{i,:}\right\|^{2}+\lambda$ and $\mu=\frac{\rho_{\min }\left(A^{\top} A\right)}{n}+\lambda$, where $\rho_{\min }$ is the smallest eigenvalue. In our experiments we set $\lambda = \frac{1}{n}$. In all plots $x$-axis is the number of communicated bits, and $y$-axis is the squared norm of difference between current iterate and solution. 

{\bf Compression operator}. In all experiments we used random sparsification as compression operator:
$$C(x)=\frac{d}{k} \sum_{i \in S} x_{i} e_{i},$$
where $S$ is a random subset of $\left\lbrace 1, 2, \ldots , d\right\rangle$ of cardinality $k$ chosen uniformly at random, and $e_i$ is the $i$-th standard unit basis vector in $\mathbb{R}^d$.

{\bf Hardware and software}. We use real datasets from
open LIBSVM corpus~\citep{chang2011libsvm} (Modified BSD License~www.csie.ntu.edu.tw/~cjlin/libsvm/) and synthetic datasets from scikit-learn.datasets~\citep{scikit-learn} (BSD License https://scikit-learn.org). We implemented all algorithms in Python. All methods were evaluated on a computer with an Intel(R) Xeon(R) Gold 6146 CPU at 3.20GHz, having 24 cores. You can find more details and additional experiments in supplementary materials. 
\begin{figure*}[t]
	\centering
	\begin{tabular}{cc}
		\includegraphics[scale=0.22]{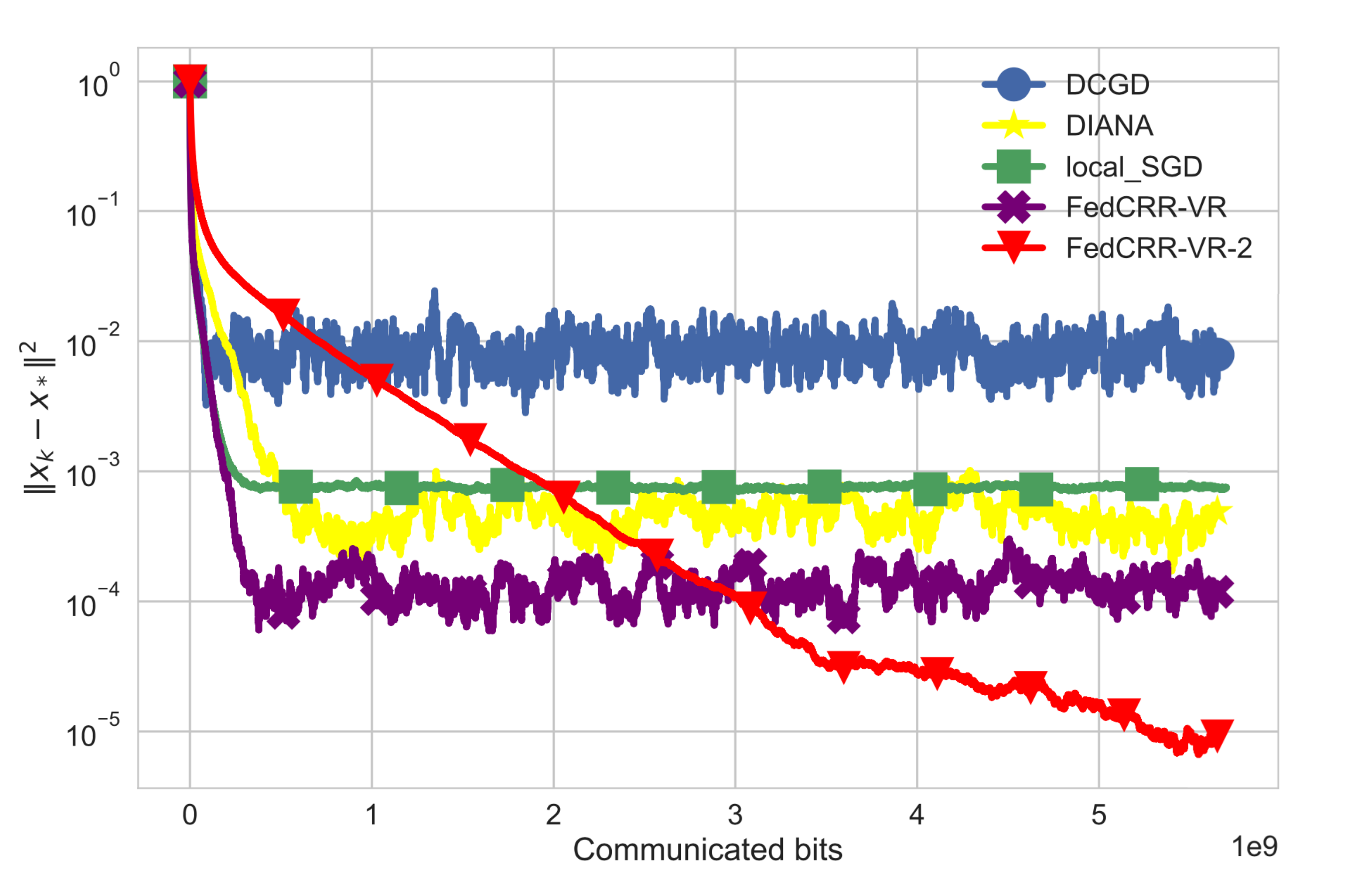}&
		\includegraphics[scale=0.18]{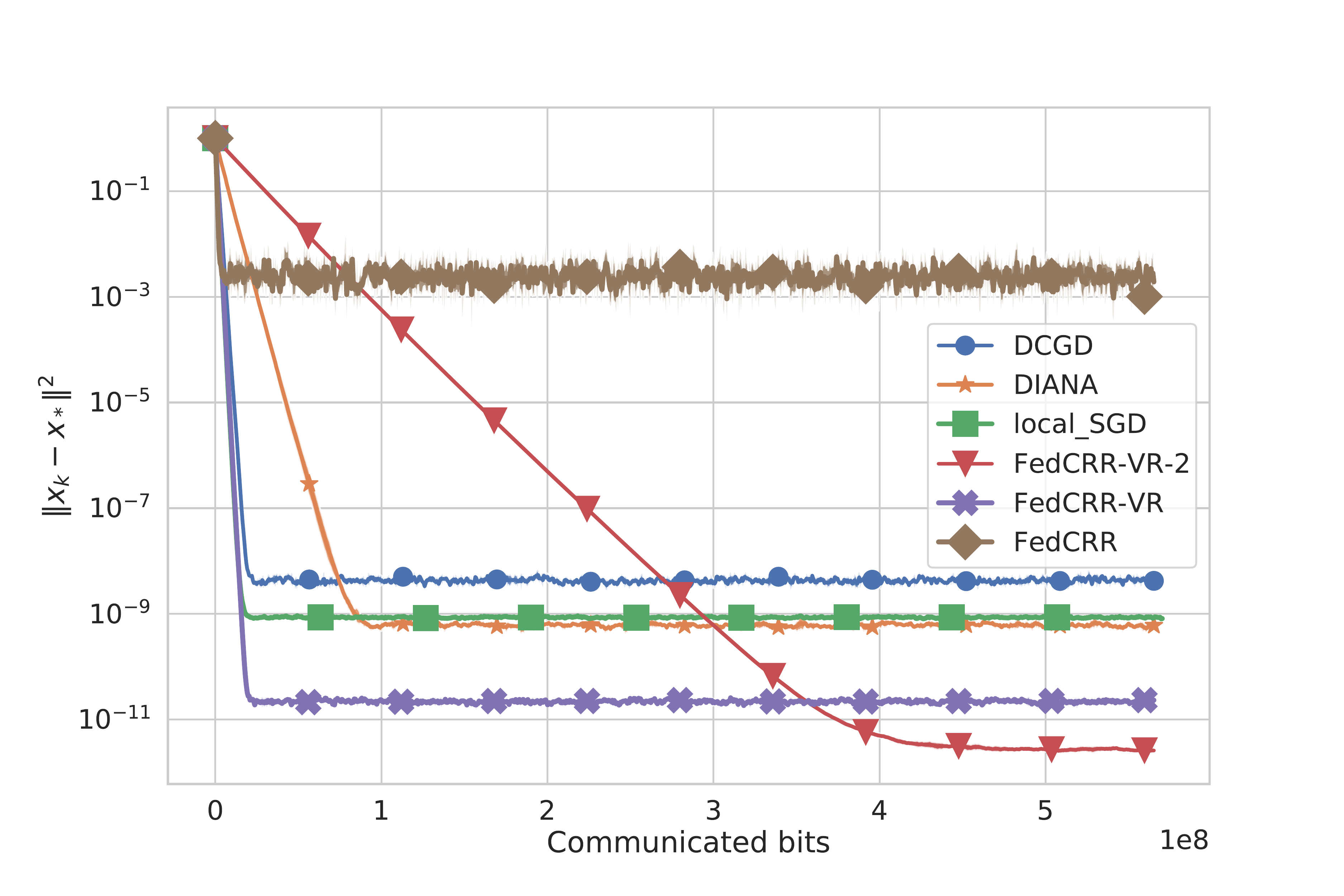}
	\end{tabular}
	\begin{tabular}{c}
		\includegraphics[scale=0.2]{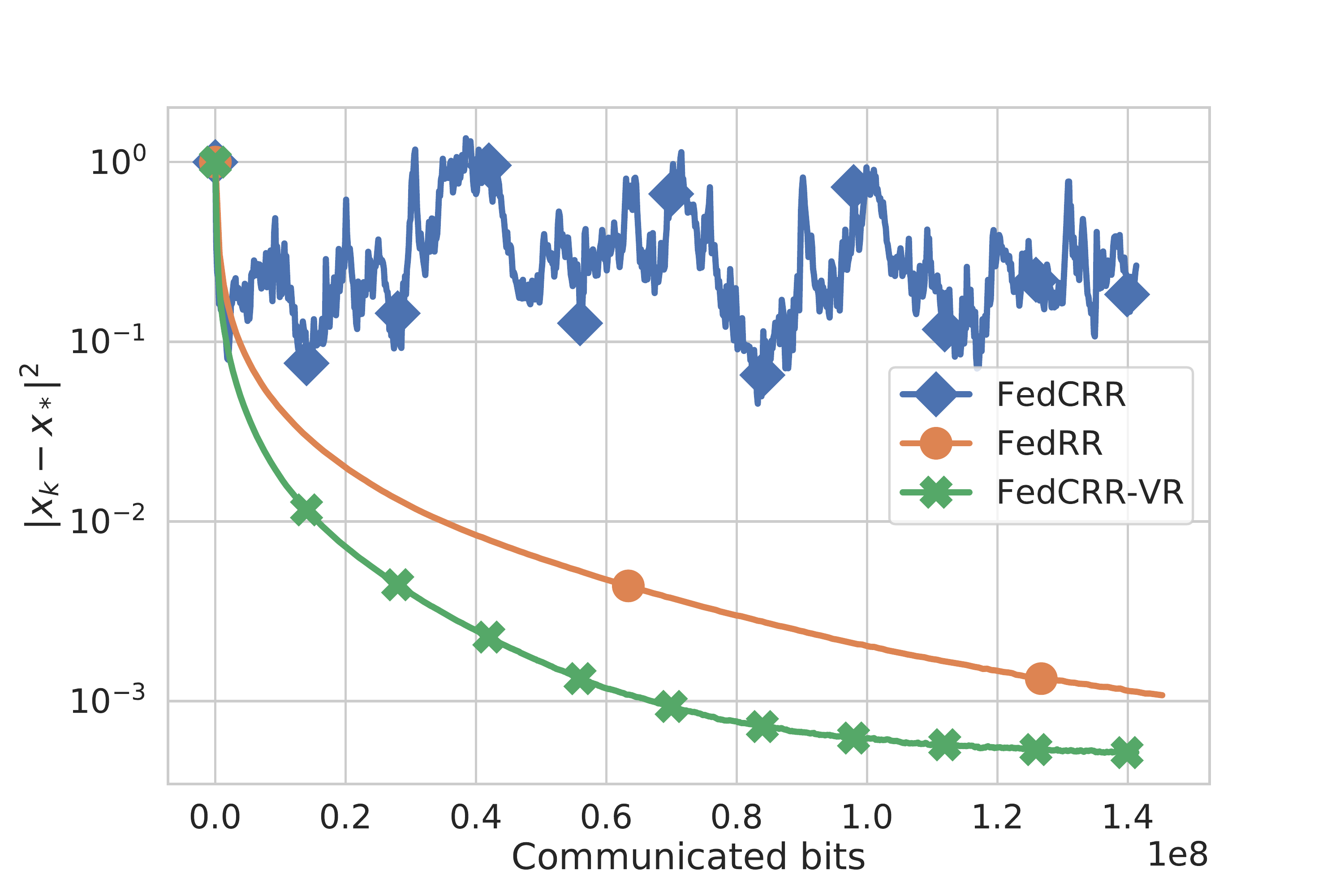}
	\end{tabular}
	\caption{Comparison of distributed methods and versions of Compressed Federated Random Reshuffling: FedCRR, FedCRR-VR, FedCRR-VR-2 on synthetic datasets with the different data points. In order to have fair competition, we used the same parameter $\omega$ for all methods with compression. }
	\label{fig:svrg_types}
\end{figure*}

{\bf Results}. We have a very tight match between our theory and the numerical results. As we can see, Compressed Federated Random Reshuffling cannot get appropriate accuracy since we have a huge compression variance term. It means that this method can be used only if the required accuracy is not high. However, we can see that variance-reduced methods show better convergence. While FedRR-VR has the same linear rate, the solution's neighborhood is smaller in comparison to other methods. FedRR-VR-2 has a slower linear rate because of stepsize requirements, but the neighborhood of the solution is the smallest and allows to get a much better solution to the problem. 

\section{Conclusion}
In this work, we propose three new algorithms: Compressed Federated Random Reshuffling and two variance-reduced variants. These methods are first-of-its-kind algorithms that include three popular approaches: periodic aggregation, compressed updates, and sampling without replacement. Moreover, we sequentially applied variance reduction mechanisms for compression and then for Random Reshuffling. We provide the first analysis under general assumptions. Experimental results confirm our theoretical findings. Thus, we gain a deeper theoretical understanding of how these algorithms work and hope that this will inspire researchers to develop further and analyze methods for Federated learning. In future work, we desire to get rid of the client drift term in the neighborhood of the solution and get an algorithm that will converge linearly to the exact solution. We also want to allow a method to have partial participation of clients since it is essential for a Federated Learning setting. We also believe that our theoretical and practical results can be applied to other aspects of machine learning and federated learning, leading to improvements in current and future applications.

\nocite{langley00}

\bibliography{biblio}

\begin{thebibliography}{48}
\providecommand{\natexlab}[1]{#1}
\providecommand{\url}[1]{\texttt{#1}}
\expandafter\ifx\csname urlstyle\endcsname\relax
  \providecommand{\doi}[1]{doi: #1}\else
  \providecommand{\doi}{doi: \begingroup \urlstyle{rm}\Url}\fi

\bibitem[Aji \& Heafield(2017)Aji and Heafield]{aji2017sparse}
Aji, A.~F. and Heafield, K.
\newblock Sparse communication for distributed gradient descent.
\newblock \emph{arXiv preprint arXiv:1704.05021}, 2017.

\bibitem[Alistarh et~al.(2017)Alistarh, Grubic, Li, Tomioka, and
  Vojnovic]{alistarh2017qsgd}
Alistarh, D., Grubic, D., Li, J., Tomioka, R., and Vojnovic, M.
\newblock Qsgd: Communication-efficient sgd via gradient quantization and
  encoding.
\newblock \emph{Advances in Neural Information Processing Systems},
  30:\penalty0 1709--1720, 2017.

\bibitem[Alistarh et~al.(2018)Alistarh, Hoefler, Johansson, Khirirat,
  Konstantinov, and Renggli]{alistarh2018convergence}
Alistarh, D., Hoefler, T., Johansson, M., Khirirat, S., Konstantinov, N., and
  Renggli, C.
\newblock The convergence of sparsified gradient methods.
\newblock \emph{arXiv preprint arXiv:1809.10505}, 2018.

\bibitem[Basu et~al.(2020)Basu, Data, Karakus, and
  Diggavi]{Basu2020QsparseLocalSGDDS}
Basu, D., Data, D., Karakus, C., and Diggavi, S.
\newblock Qsparse-local-sgd: Distributed sgd with quantization, sparsification,
  and local computations.
\newblock \emph{IEEE Journal on Selected Areas in Information Theory},
  1:\penalty0 217--226, 2020.

\bibitem[Bengio(2012)]{Bengio2012}
Bengio, Y.
\newblock {Practical recommendations for gradient-based training of deep
  architectures}.
\newblock \emph{Neural Networks: Tricks of the Trade}, pp.\  437–478, 2012.
\newblock ISSN 1611-3349.
\newblock \doi{10.1007/978-3-642-35289-8_26}.

\bibitem[Bernstein et~al.(2018)Bernstein, Wang, Azizzadenesheli, and
  Anandkumar]{bernstein2018signsgd}
Bernstein, J., Wang, Y.-X., Azizzadenesheli, K., and Anandkumar, A.
\newblock signsgd: Compressed optimisation for non-convex problems.
\newblock In \emph{International Conference on Machine Learning}, pp.\
  560--569. PMLR, 2018.

\bibitem[Bottou(2009)]{bottou2009curiously}
Bottou, L.
\newblock Curiously fast convergence of some stochastic gradient descent
  algorithms.
\newblock 2009.

\bibitem[Chang \& Lin(2011)Chang and Lin]{chang2011libsvm}
Chang, C.-C. and Lin, C.-J.
\newblock Libsvm: a library for support vector machines.
\newblock \emph{ACM Transactions on Intelligent Systems and Technology (TIST)},
  2\penalty0 (3):\penalty0 1--27, 2011.

\bibitem[Chraibi et~al.(2019)Chraibi, Khaled, Kovalev, Richt{\'a}rik, Salim,
  and Tak{\'a}{\v{c}}]{chraibi2019distributed}
Chraibi, S., Khaled, A., Kovalev, D., Richt{\'a}rik, P., Salim, A., and
  Tak{\'a}{\v{c}}, M.
\newblock Distributed fixed point methods with compressed iterates.
\newblock \emph{arXiv preprint arXiv:1912.09925}, 2019.

\bibitem[Defazio et~al.(2014{\natexlab{a}})Defazio, Bach, and
  Lacoste-Julien]{defazio2014saga}
Defazio, A., Bach, F., and Lacoste-Julien, S.
\newblock {SAGA} : A fast incremental gradient method with support for
  non-strongly convex composite objectives.
\newblock \emph{arXiv preprint arXiv:1407.0202}, 2014{\natexlab{a}}.

\bibitem[Defazio et~al.(2014{\natexlab{b}})Defazio, Domke, and
  Caetano]{pmlr-v32-defazio14}
Defazio, A., Domke, J., and Caetano.
\newblock Finito: A faster, permutable incremental gradient method for big data
  problems.
\newblock In Xing, E.~P. and Jebara, T. (eds.), \emph{Proceedings of the 31st
  International Conference on Machine Learning}, volume~32 of \emph{Proceedings
  of Machine Learning Research}, pp.\  1125--1133, Bejing, China, 22--24 Jun
  2014{\natexlab{b}}. PMLR.

\bibitem[Gorbunov et~al.(2020)Gorbunov, Hanzely, and
  Richt{\'a}rik]{gorbunov2020unified}
Gorbunov, E., Hanzely, F., and Richt{\'a}rik, P.
\newblock A unified theory of sgd: Variance reduction, sampling, quantization
  and coordinate descent.
\newblock In \emph{International Conference on Artificial Intelligence and
  Statistics}, pp.\  680--690. PMLR, 2020.

\bibitem[Gower et~al.(2019)Gower, Loizou, Qian, Sailanbayev, Shulgin, and
  Richt{\'a}rik]{gower2019sgd}
Gower, R.~M., Loizou, N., Qian, X., Sailanbayev, A., Shulgin, E., and
  Richt{\'a}rik, P.
\newblock {SGD}: {G}eneral analysis and improved rates.
\newblock In \emph{International Conference on Machine Learning}, pp.\
  5200--5209. PMLR, 2019.

\bibitem[G{\"u}rb{\"u}zbalaban et~al.(2019)G{\"u}rb{\"u}zbalaban, Ozdaglar, and
  Parrilo]{gurbuzbalaban2019random}
G{\"u}rb{\"u}zbalaban, M., Ozdaglar, A., and Parrilo, P.~A.
\newblock Why random reshuffling beats stochastic gradient descent.
\newblock \emph{Mathematical Programming}, pp.\  1--36, 2019.

\bibitem[Haddadpour et~al.(2021)Haddadpour, Kamani, Mokhtari, and
  Mahdavi]{haddadpour2021federated}
Haddadpour, F., Kamani, M.~M., Mokhtari, A., and Mahdavi, M.
\newblock Federated learning with compression: Unified analysis and sharp
  guarantees.
\newblock In \emph{International Conference on Artificial Intelligence and
  Statistics}, pp.\  2350--2358. PMLR, 2021.

\bibitem[Haochen \& Sra(2019)Haochen and Sra]{haochen2018random}
Haochen, J. and Sra, S.
\newblock {Random shuffling beats {SGD} after finite epochs}.
\newblock In Chaudhuri, K. and Salakhutdinov, R. (eds.), \emph{Proceedings of
  the 36th International Conference on Machine Learning}, volume~97 of
  \emph{Proceedings of Machine Learning Research}, pp.\  2624--2633, Long
  Beach, California, USA, 09--15 Jun 2019. PMLR.

\bibitem[Horvath et~al.(2019)Horvath, Ho, Horvath, Sahu, Canini, and
  Richtarik]{horvath2019natural}
Horvath, S., Ho, C.-Y., Horvath, L., Sahu, A.~N., Canini, M., and Richtarik, P.
\newblock Natural compression for distributed deep learning.
\newblock \emph{arXiv preprint arXiv:1905.10988}, 2019.

\bibitem[Johnson \& Zhang(2013)Johnson and Zhang]{johnson2013accelerating}
Johnson, R. and Zhang, T.
\newblock Accelerating stochastic gradient descent using predictive variance
  reduction.
\newblock \emph{Advances in Neural Information Processing Systems},
  26:\penalty0 315--323, 2013.

\bibitem[Karimireddy et~al.(2019)Karimireddy, Kale, Mohri, Reddi, Stich, and
  Suresh]{Karimireddy2019}
Karimireddy, S.~P., Kale, S., Mohri, M., Reddi, S.~J., Stich, S.~U., and
  Suresh, A.~T.
\newblock {SCAFFOLD: Stochastic Controlled Averaging for On-Device Federated
  Learning}.
\newblock \emph{arXiv preprint arXiv:1910.06378}, 2019.

\bibitem[Khaled \& Richt{\'a}rik(2019)Khaled and
  Richt{\'a}rik]{khaled2019gradient}
Khaled, A. and Richt{\'a}rik, P.
\newblock Gradient descent with compressed iterates.
\newblock \emph{arXiv preprint arXiv:1909.04716}, 2019.

\bibitem[Khaled et~al.(2019)Khaled, Mishchenko, and
  Richt{\'a}rik]{khaled2019first}
Khaled, A., Mishchenko, K., and Richt{\'a}rik, P.
\newblock First analysis of local gd on heterogeneous data.
\newblock \emph{arXiv preprint arXiv:1909.04715}, 2019.

\bibitem[Khaled et~al.(2020)Khaled, Mishchenko, and
  Richt{\'a}rik]{khaled2020tighter}
Khaled, A., Mishchenko, K., and Richt{\'a}rik, P.
\newblock Tighter theory for local sgd on identical and heterogeneous data.
\newblock In \emph{International Conference on Artificial Intelligence and
  Statistics}, pp.\  4519--4529. PMLR, 2020.

\bibitem[Kone{\v{c}}n{\`y} et~al.(2016)Kone{\v{c}}n{\`y}, McMahan, Yu,
  Richt{\'a}rik, Suresh, and Bacon]{konevcny2016federated}
Kone{\v{c}}n{\`y}, J., McMahan, H.~B., Yu, F.~X., Richt{\'a}rik, P., Suresh,
  A.~T., and Bacon, D.
\newblock Federated learning: Strategies for improving communication
  efficiency.
\newblock \emph{arXiv preprint arXiv:1610.05492}, 2016.

\bibitem[Kovalev et~al.(2020)Kovalev, Horv{\'a}th, and
  Richt{\'a}rik]{kovalev2020don}
Kovalev, D., Horv{\'a}th, S., and Richt{\'a}rik, P.
\newblock Don’t jump through hoops and remove those loops: {SVRG} and
  katyusha are better without the outer loop.
\newblock In \emph{Algorithmic Learning Theory}, pp.\  451--467. PMLR, 2020.

\bibitem[Lin et~al.(2018)Lin, Stich, Patel, and Jaggi]{lin2018don}
Lin, T., Stich, S.~U., Patel, K.~K., and Jaggi, M.
\newblock Don't use large mini-batches, use local sgd.
\newblock \emph{arXiv preprint arXiv:1808.07217}, 2018.

\bibitem[Lin et~al.(2017)Lin, Han, Mao, Wang, and Dally]{lin2017deep}
Lin, Y., Han, S., Mao, H., Wang, Y., and Dally, W.~J.
\newblock Deep gradient compression: Reducing the communication bandwidth for
  distributed training.
\newblock \emph{arXiv preprint arXiv:1712.01887}, 2017.

\bibitem[Malinovsky et~al.(2021)Malinovsky, Sailanbayev, and
  Richt{\'a}rik]{malinovsky2021random}
Malinovsky, G., Sailanbayev, A., and Richt{\'a}rik, P.
\newblock Random reshuffling with variance reduction: New analysis and better
  rates.
\newblock \emph{arXiv preprint arXiv:2104.09342}, 2021.

\bibitem[McMahan et~al.(2017)McMahan, Moore, Ramage, Hampson, and
  y~Arcas]{mcmahan2017communication}
McMahan, B., Moore, E., Ramage, D., Hampson, S., and y~Arcas, B.~A.
\newblock Communication-efficient learning of deep networks from decentralized
  data.
\newblock In \emph{Artificial Intelligence and Statistics}, pp.\  1273--1282.
  PMLR, 2017.

\bibitem[Mishchenko et~al.(2019)Mishchenko, Gorbunov, Tak{\'a}{\v{c}}, and
  Richt{\'a}rik]{mishchenko2019distributed}
Mishchenko, K., Gorbunov, E., Tak{\'a}{\v{c}}, M., and Richt{\'a}rik, P.
\newblock Distributed learning with compressed gradient differences.
\newblock \emph{arXiv preprint arXiv:1901.09269}, 2019.

\bibitem[Mishchenko et~al.(2020)Mishchenko, Khaled, and
  Richt{\'a}rik]{mishchenko2020random}
Mishchenko, K., Khaled, A., and Richt{\'a}rik, P.
\newblock Random reshuffling: Simple analysis with vast improvements.
\newblock \emph{Advances in Neural Information Processing Systems}, 33, 2020.

\bibitem[Mishchenko et~al.(2021)Mishchenko, Khaled, and
  Richt{\'a}rik]{mishchenko2021proximal}
Mishchenko, K., Khaled, A., and Richt{\'a}rik, P.
\newblock Proximal and federated random reshuffling.
\newblock \emph{arXiv preprint arXiv:2102.06704}, 2021.

\bibitem[Mokhtari et~al.(2018)Mokhtari, G\"{u}rb\"{u}zbalaban, and
  Ribeiro]{mokhtari2018surpassing}
Mokhtari, A., G\"{u}rb\"{u}zbalaban, M., and Ribeiro, A.
\newblock Surpassing gradient descent provably: A cyclic incremental method
  with linear convergence rate.
\newblock \emph{SIAM Journal on Optimization}, 28\penalty0 (2):\penalty0
  1420--1447, 2018.

\bibitem[Nagaraj et~al.(2019)Nagaraj, Jain, and Netrapalli]{Nagaraj2019}
Nagaraj, D., Jain, P., and Netrapalli, P.
\newblock {SGD without replacement: sharper rates for general smooth convex
  functions}.
\newblock In Chaudhuri, K. and Salakhutdinov, R. (eds.), \emph{Proceedings of
  the 36th International Conference on Machine Learning}, volume~97 of
  \emph{Proceedings of Machine Learning Research}, pp.\  4703--4711, Long
  Beach, California, USA, 09--15 Jun 2019. PMLR.

\bibitem[Park \& Ryu(2020)Park and Ryu]{park2020linear}
Park, Y. and Ryu, E.~K.
\newblock Linear convergence of cyclic {SAGA}.
\newblock \emph{Optimization Letters}, 14\penalty0 (6):\penalty0 1583--1598,
  2020.

\bibitem[Pedregosa et~al.(2011)Pedregosa, Varoquaux, Gramfort, Michel, Thirion,
  Grisel, Blondel, Prettenhofer, Weiss, Dubourg, Vanderplas, Passos,
  Cournapeau, Brucher, Perrot, and Duchesnay]{scikit-learn}
Pedregosa, F., Varoquaux, G., Gramfort, A., Michel, V., Thirion, B., Grisel,
  O., Blondel, M., Prettenhofer, P., Weiss, R., Dubourg, V., Vanderplas, J.,
  Passos, A., Cournapeau, D., Brucher, M., Perrot, M., and Duchesnay, E.
\newblock Scikit-learn: Machine learning in {P}ython.
\newblock \emph{Journal of Machine Learning Research}, 12:\penalty0 2825--2830,
  2011.

\bibitem[Rajput et~al.(2020)Rajput, Gupta, and Papailiopoulos]{Rajput2020}
Rajput, S., Gupta, A., and Papailiopoulos, D.
\newblock {Closing the convergence gap of SGD without replacement}.
\newblock \emph{arXiv preprint arXiv:2002.10400}, 2020.

\bibitem[Ramezani-Kebrya et~al.(2019)Ramezani-Kebrya, Faghri, and
  Roy]{ramezani2019nuqsgd}
Ramezani-Kebrya, A., Faghri, F., and Roy, D.~M.
\newblock Nuqsgd: Improved communication efficiency for data-parallel sgd via
  nonuniform quantization.
\newblock \emph{arXiv preprint arXiv:1908.06077}, 2019.

\bibitem[Recht \& R{\'e}(2012)Recht and R{\'e}]{recht2012toward}
Recht, B. and R{\'e}, C.
\newblock Toward a noncommutative arithmetic-geometric mean inequality:
  conjectures, case-studies, and consequences.
\newblock In \emph{Conference on Learning Theory}, pp.\  11--1. JMLR Workshop
  and Conference Proceedings, 2012.

\bibitem[Recht \& R{\'e}(2013)Recht and R{\'e}]{recht2013parallel}
Recht, B. and R{\'e}, C.
\newblock Parallel stochastic gradient algorithms for large-scale matrix
  completion.
\newblock \emph{Mathematical Programming Computation}, 5\penalty0 (2):\penalty0
  201--226, 2013.

\bibitem[Robbins \& Monro(1951)Robbins and Monro]{robbins1951stochastic}
Robbins, H. and Monro, S.
\newblock A stochastic approximation method.
\newblock \emph{The annals of mathematical statistics}, pp.\  400--407, 1951.

\bibitem[Roux et~al.(2012)Roux, Schmidt, and Bach]{roux2012stochastic}
Roux, N.~L., Schmidt, M., and Bach, F.
\newblock A stochastic gradient method with an exponential convergence rate for
  finite training sets.
\newblock \emph{arXiv preprint arXiv:1202.6258}, 2012.

\bibitem[Safran \& Shamir(2020)Safran and Shamir]{safran2020good}
Safran, I. and Shamir, O.
\newblock How good is {SGD} with random shuffling?
\newblock In \emph{Conference on Learning Theory}, pp.\  3250--3284. PMLR,
  2020.

\bibitem[Shamir et~al.(2014)Shamir, Srebro, and Zhang]{shamir2014communication}
Shamir, O., Srebro, N., and Zhang, T.
\newblock Communication-efficient distributed optimization using an approximate
  newton-type method.
\newblock In \emph{International conference on machine learning}, pp.\
  1000--1008. PMLR, 2014.

\bibitem[Stich(2018)]{stich2018local}
Stich, S.~U.
\newblock Local sgd converges fast and communicates little.
\newblock \emph{arXiv preprint arXiv:1805.09767}, 2018.

\bibitem[Vogels et~al.(2019)Vogels, Karimireddy, and Jaggi]{vogels2019powersgd}
Vogels, T., Karimireddy, S.~P., and Jaggi, M.
\newblock Powersgd: Practical low-rank gradient compression for distributed
  optimization.
\newblock \emph{arXiv preprint arXiv:1905.13727}, 2019.

\bibitem[Wangni et~al.(2017)Wangni, Wang, Liu, and Zhang]{wangni2017gradient}
Wangni, J., Wang, J., Liu, J., and Zhang, T.
\newblock Gradient sparsification for communication-efficient distributed
  optimization.
\newblock \emph{arXiv preprint arXiv:1710.09854}, 2017.

\bibitem[Wu et~al.(2018)Wu, Huang, Huang, and Zhang]{wu2018error}
Wu, J., Huang, W., Huang, J., and Zhang, T.
\newblock Error compensated quantized sgd and its applications to large-scale
  distributed optimization.
\newblock In \emph{International Conference on Machine Learning}, pp.\
  5325--5333. PMLR, 2018.

\bibitem[Ying et~al.(2019)Ying, Yuan, Vlaski, and Sayed]{Ying2019}
Ying, B., Yuan, K., Vlaski, S., and Sayed, A.~H.
\newblock Stochastic learning under random reshuffling with constant
  step-sizes.
\newblock In \emph{IEEE Transactions on Signal Processing}, volume~67, pp.\
  474--489, 2019.

\end{thebibliography}
\bibliographystyle{icml2021}
	\clearpage

\onecolumn
\appendix

\part*{Supplementary materials}
\section{Basic Facts}\label{seca1}
\begin{proposition}
	Let $f : \mathbb{R}^d \to \mathbb{R}$ be continuously differentiable and let $L\geq 0$. Then the following statements are equivalent:
	\begin{itemize}
		\item $f$ is $L$-smooth
		\item $2 D_{f}(x, y) \leq L\|x-y\|^{2} \text { for all } x, y \in \mathbb{R}^{d}$
		\item $\langle\nabla f(x)-\nabla f(y), x-y\rangle \leq L\|x-y\|^{2} \text { for all } x, y \in \mathbb{R}^{d}$
	\end{itemize}
\end{proposition}

\begin{proposition}
	Let $f : \mathbb{R}^d \to \mathbb{R}$ be continuously differentiable and let $\mu\geq 0$. Then the following statements are equivalent:
	\begin{itemize}
		\item $f$ is $\mu$-strongly convex
		\item $2 D_{f}(x, y) \geq \mu\|x-y\|^{2} \text { for all } x, y \in \mathbb{R}^{d}$
		\item $\langle\nabla f(x)-\nabla f(y), x-y\rangle \geq \mu\|x-y\|^{2} \text { for all } x, y \in \mathbb{R}^{d}$
	\end{itemize}
\end{proposition}
Note that the $\mu = 0$ case reduces to convexity.

\begin{proposition}
	Let $f : \mathbb{R}^d \to \mathbb{R}$ be continuously differentiable and $L > 0$. Then the following statements are equivalent:
	\begin{itemize}
		\item $f$ is convex and $L$-smooth
		\item $0 \leq 2 D_{f}(x, y) \leq L\|x-y\|^{2} \text { for all } x, y \in \mathbb{R}^{d}$
		\item $\frac{1}{L}\|\nabla f(x)-\nabla f(y)\|^{2} \leq 2 D_{f}(x, y) \text { for all } x, y \in \mathbb{R}^{d}$
		\item $\frac{1}{L}\|\nabla f(x)-\nabla f(y)\|^{2} \leq\langle\nabla f(x)-\nabla f(y), x-y\rangle \text { for all } x, y \in \mathbb{R}^{d}$
	\end{itemize}
\end{proposition}

\begin{proposition}[Jensen's inequality]
	Let $f: \mathbb{R}^{d} \to \mathbb{R}$ be a convex function, $x_1,\ldots,x_m \in \mathbb{R}^{d}$ and $\lambda_1,\ldots, \lambda_m$ be nonnegative real numbers adding up to 1. Then  
	$$f\left(\sum_{i=1}^{m} \lambda_{i} x_{i}\right) \leq \sum_{i=1}^{m} \lambda_{i} f\left(x_{i}\right)$$
\end{proposition}

\begin{proposition}
	For all $a, b \in \mathbb{R}^{d}$  and $t > 0$ the following inequalities holds:
	\begin{align*}
	\langle a, b\rangle &\leq \frac{\|a\|^{2}}{2 t}+\frac{t\|b\|^{2}}{2}\\
	\|a+b\|^{2} &\leq 2\|a\|^{2}+2\|b\|^{2}\\
	\frac{1}{2}\|a\|^{2}-\|b\|^{2} &\leq\|a+b\|^{2}
	\end{align*}
	
\end{proposition}
\section{General lemmas}
\subsection{Proposition 1}
We need to prove a basic fact which will be used later.
\begin{proposition}
	Let us consider 
	$$x^n_{*,m} = x_* - \gamma\sum_{i=0}^{n-1}\nabla f_{\pi_i,m}(x_{*}),$$
	then 
	$$\frac{1}{M}\sum_{m=1}^{M} x^n_{*,m} = x_*.$$
	\begin{proof}
		We start from the definition:
		\begin{align*}
		\frac{1}{M}\sum_{m=1}^{M} x^n_{*,m} &= \frac{1}{M}\sum_{m=1}^{M}\left(x_* - \gamma\sum_{i=0}^{n-1}\nabla f_{\pi_i,m}(x_{*})\right)\\
		&=\frac{1}{M}\sum_{m=1}^{M} x_* - \frac{1}{M}\sum_{m=1}^{M}\sum_{i=0}^{n-1}\nabla f_{\pi_i,m}(x_*)\\
		& = x_* - \nabla f(x_*)\\
		&=x_*.
		\end{align*}
	\end{proof}
	
\end{proposition}

\subsection{Proof of Theorem~\ref{thm:rad}}
For completeness we include the proof of important theorem introduced in~\citet{mishchenko2020random}.
\begin{proof}
	\begin{align*}
	\mathbb{E}\left[D_{f_{\pi_{i}}}\left(x_{*}^{i}, x_{*}\right)\right] \leq \mathbb{E}\left[\frac{L}{2}\left\|x_{*}^{i}-x_{*}\right\|^{2}\right] & \leq \frac{L_{\max }}{2} \mathbb{E}\left[\left\|x_{*}^{i}-x_{*}\right\|^{2}\right] \\
	& = \frac{\gamma^{2} L_{\max }}{2} \mathbb{E}\left[\left\|\sum_{j=0}^{i-1} \nabla f_{\pi_{j}}\left(x_{*}\right)\right\|^{2}\right] \\
	&=\frac{\gamma^{2} L_{\max } i^{2}}{2} \mathbb{E}\left[\left\|\frac{1}{i} \sum_{j=0}^{i-1} \nabla f_{\pi_{j}}\left(x_{*}\right)\right\|^{2}\right] \\
	&=\frac{\gamma^{2} L_{\max } i^{2}}{2} \mathbb{E}\left[\left\|\bar{X}_{\pi}\right\|^{2}\right],
	\end{align*}
	where $\bar{X}_{\pi}=\frac{1}{j} \sum_{j=0}^{i-1} X_{\pi_{j}} \text { with } X_{j} \stackrel{\text { def }}{=} \nabla f_{j}\left(x_{*}\right) \text { for } j=1,2, \ldots, n \text { . Since } \bar{X}=\nabla f\left(x_{*}\right),$ by applying Lemma 1 in \citep{mishchenko2020random}.
	$$\mathbb{E}\left[\left\|\bar{X}_{\pi}\right\|^{2}\right]=\|\bar{X}\|^{2}+\mathbb{E}\left[\left\|\bar{X}_{\pi}-\bar{X}\right\|^{2}\right] =\left\|\nabla f\left(x_{*}\right)\right\|^{2}+\frac{n-i}{i(n-1)} \sigma_{*}^{2} .$$ It remains to combine both terms and use the bounds $i^2 \leq n^2$ and $i(n-i) \leq \frac{n(n-1)}{2}$, which holds for all $i \in \left\lbrace1, 2, \ldots , n-1 \right\rbrace$, and divide both sides of the resulting inequality by $\gamma^2$.
\end{proof}

\subsection{Proof of Lemma~\ref{lemma:rad} }	
\begin{proof}
	We start from Theorem~\ref{thm:rad}. Then for reformulated problem we have 
	$$n \sigma_{*}^{2} \stackrel{\text { def }}{=} \sum_{i=1}^{n}\left\|\nabla f_{i}\left(\boldsymbol{x}_{*}\right)-\nabla f\left(\boldsymbol{x}_{*}\right)\right\|^{2}=\sum_{i=1}^{n} \sum_{m=1}^{M}\left\|\nabla f_{m i}\left(x_{*}\right)-\frac{1}{n} \nabla F_{m}\left(x_{*}\right)\right\|^{2}.$$
	For inner sum we have a bound from~\citet{mishchenko2021proximal}:
	$$
	\sum_{i=1}^{n}\left\|\nabla f_{m i}\left(x_{*}\right)-\frac{1}{n} \nabla F_{m}\left(x_{*}\right)\right\|^{2}
	\leq  n \sigma_{m, *}^{2}+\left\|\nabla F_{m}\left(x_{*}\right)\right\|^{2}.$$
	Also, we have 
	$$n^{2}\left\|\nabla f\left(\boldsymbol{x}_{*}\right)\right\|^{2}=n^{2}\left\|\frac{1}{n} \sum_{i=1}^{n} \nabla f_{i}\left(\boldsymbol{x}_{*}\right)\right\|^{2}=\sum_{m=1}^{M}\left\|\sum_{i=1}^{n} \nabla f_{m i}\left(x_{*}\right)\right\|^{2}=\sum_{m=1}^{M}\left\|\nabla F_{m}\left(x_{*}\right)\right\|^{2}.$$
	Plugging the last two inequalities back inside the first bound on $\sigma_{\mathrm{rad}}^{2}$, we get the lemma’s statement.
\end{proof}

\section{Analysis of Algorithm~\ref{alg:fed_rr}}
\subsection{Proof of Theorem 1}
\begin{proof}
	We start from conditional expectation
	\begin{align*}
	\mathbb{E} \left[\|x_{t+1} - x_*\|^2\mid x^n_{t,m}\right] &= 
	\mathbb{E}\left[\left\|\frac{1}{M}\sum_{m=1}^{M}C(x^n_{t,m})-x_*\right\|^2\bigg| x^n_{t,m}\right]\\
	&\leq \mathbb{E}\left\|\frac{1}{M}\sum_{m=1}^{M}C(x^n_{t,m}) - \frac{1}{M}\sum_{m=1}^{M}x^n_{t,m}\bigg|x^n_{t,m}\right\|^2+\left\| \frac{1}{M}\sum_{m=1}^{M}x^n_{t,m} - x_* \right\|^2\\
	&\leq \frac{\omega}{M^2}\sum_{m=1}^{M}\|x^n_{t,m}\|^2+\frac{1}{M}\sum_{m=1}^{M}\|x^n_{t,m}- x^n_{*,m}\|^2\\
	&\leq \frac{2\omega}{M^2}\sum_{m=1}^{M}\|x^n_{t,m} - x^n_{*,m}\|^2+\frac{1}{M}\sum_{m=1}^{M}\|x^n_{t,m}- x^n_{*,m}\|^2 + \frac{2\omega}{M^2} \sum_{m=1}^{M}\|x^n_{*,m}\|^2\\
	&\leq (1-\gamma\mu)^n\left(1+\frac{2\omega}{M}\right)\|x_t-x_*\|^2+\frac{2\omega}{M}\frac{1}{M}\sum_{m=1}^{M}\|x^n_{*,m}\|^2\\
	&+2\left(1+\frac{2\omega}{M}\right)\gamma^3\sigma^2_{rad}\left(\sum_{j=0}^{n-1}(1-\gamma\mu)^j\right).
	\end{align*}
	Using tower property we get
	$$\mathbb{E} \left[\|x_{t+1} - x_*\|^2\right] =\mathbb{E} \left[\mathbb{E} \left[\|x_{t+1} - x_*\|^2\mid x^n_{t,m}\right] \right]. $$
	Utilizing this property we have 
	\begin{align*}
	\mathbb{E} \left[\|x_{t+1} - x_*\|^2\right] &\leq (1-\gamma\mu)^n\left(1+\frac{2\omega}{M}\right)\mathbb{E}\left[\|x_t-x_*\|^2\right]+\frac{2\omega}{M}\frac{1}{M}\sum_{m=1}^{M}\mathbb{E}\|x^n_{*,m}\|^2\\
	&+2\left(1+\frac{2\omega}{M}\right)\gamma^3\sigma^2_{rad}\left(\sum_{j=0}^{n-1}(1-\gamma\mu)^j\right).
	\end{align*}
	Unrolling this recursion we get
	\begin{align*}
	\mathbb{E} \left[\|x_{t+1} - x_*\|^2\right] &\leq \left((1-\gamma\mu)^n\left(1+\frac{2\omega}{M}\right)\right)^T\|x_0-x_*\|^2\\
	&+\sum_{i=0}^{T-1} \left((1-\gamma\mu)^n\left(1+\frac{2\omega}{M}\right)\right)^i \frac{2\omega}{M}\frac{1}{M}\sum_{m=1}^{M}\mathbb{E}\|x^n_{*,m}\|^2\\
	&+2\sum_{i=0}^{T-1} \left((1-\gamma\mu)^n\left(1+\frac{2\omega}{M}\right)\right)^i\left(1+\frac{2\omega}{M}\right)\gamma^3\sigma^2_{rad}\left(\sum_{j=0}^{n-1}(1-\gamma\mu)^j\right).
	\end{align*}
	Using assumption of compression operator we have
	$$(1-\gamma\mu)^n\left(1+\frac{2\omega}{M}\right)\leq (1-\gamma\mu)^{\frac{n}{2}}.$$
	Also let us look at last term:
	\begin{align*}
	\left(1+\frac{2\omega}{M}\right)\left(\sum_{i=0}^{T-1}(1-\gamma\mu)^i\right)&\leq \sum_{i=0}^{T-1}(1-\gamma\mu)^i\left(1+\frac{2\omega}{M}\right)^i\\
	&\leq \sum_{j=0}^{T-1}\left((1-\gamma\mu)\left(1+\frac{2\omega}{M}\right)\right)^i\\
	&\leq \sum_{i=0}^{T-1}(1-\gamma\mu)^\frac{nj}{2}.
	\end{align*}
	Moreover, we have this bound for geometric sequence:
	\begin{align*}
	\left(\sum_{i=0}^{T-1}(1-\gamma\mu)^\frac{nj}{2} \right)\left(\sum_{j=0}^{n-1}(1-\gamma\mu)^j\right)\leq \sum_{i=0}^{T-1}\sum_{j=0}^{n-1}(1-\gamma\mu)^{\frac{ni}{2}+j} \leq \frac{1}{\gamma\mu}.
	\end{align*} 
	The same bound we have for the second sum:
	$$\left(\sum_{i=0}^{T-1}(1-\gamma\mu)^\frac{nj}{2} \right)\leq \frac{1}{\gamma\mu}.$$
	Finally, we have the following:
	\begin{align*}
	\mathbb{E} \left[\|x_{t+1} - x_*\|^2\right] \leq (1-\gamma\mu)^{\frac{nT}{2}}\|x_{0} - x_*\|^2+\frac{2}{\mu}\gamma^2\sigma^2_{rad}+\frac{2\omega}{M}\frac{1}{\gamma\mu}\frac{1}{M}\sum_{m=1}^{M}\mathbb{E}\|x^n_{*,m}\|^2.
	\end{align*}
	Using Lemma we have 
	\begin{align*}
	\mathbb{E} \left[\|x_{t+1} - x_*\|^2\right] &\leq (1-\gamma\mu)^{\frac{nT}{2}}\|x_{0} - x_*\|^2+\frac{2}{\mu}\gamma^2L_{\max}\frac{1}{M}\sum_{m=1}^{M}\left(\|F_m(x_*)\|^2+\frac{n}{4}\sigma_{*,m}^2\right)\\
	&+\frac{2\omega}{M}\frac{1}{\gamma\mu}\frac{1}{M}\sum_{m=1}^{M}\mathbb{E}\|x^n_{*,m}\|^2.
	\end{align*}
\end{proof}

\section{Analysis of Algorithm~\ref{alg:vr_fed_rr} and Algorithm~\ref{alg:dvr_fed_rr}}

\subsection{Lemma 2}
For Algorithm~\ref{alg:vr_fed_rr} and Algorithm~\ref{alg:dvr_fed_rr} the following inequality holds:
$$ \mathbb{E}\left[\| x_{t+1} - x_*\|^2\mid x_t,h_{t,m}\right] \leq  \frac{\eta^2}{M^2}\omega \sum_{m=1}^{M} \|x^n_{t,m} - h_{t,m}\|^2+(1-\eta)\|x_t-x_*\|^2+\eta\frac{1}{M}\sum_{m=1}^{M}\|x^n_{t,m} - x^n_{*,m}\|^2$$
\begin{proof}
	Let us use property of compression operator:
	\begin{align*}
	\mathbb{E}\left[\| x_{t+1} - x_*\|^2\mid x_t,h_{t,m}\right] &= \mathbb{E}\left[\left\| (1-\eta)x_t+\eta\frac{1}{M}\sum_{m=1}^{M} \left( C(x^n_{t,m} - h_{t,m})+h_{t,m} \right) \right\|^2\Bigg| x_t,h_{t,m}\right]\\
	&=\mathbb{E}\left[\left\|\eta\frac{1}{M}\sum_{m=1}^{M} \mathcal{C}(x^n_{t,m} - h_{t,m}) - \eta \frac{1}{M}\sum_{m=1}^{M}(x^n_{t,m} - h_{t,m}) \right\|^2\Bigg|x_t,h_{t,m}\right]\\
	&+\left\| (1-\eta)x_t+\eta\frac{1}{M}\sum_{m=1}^{M}x^n_{t,m} - x_* \right\|^2\\
	&\leq \frac{\eta^2}{M^2}\omega \sum_{m=1}^{M} \|x^n_{t,m} - h_{t,m}\|^2+(1-\eta)\|x_t-x_*\|^2+\eta\frac{1}{M}\sum_{m=1}^{M}\|x^n_{t,m} - x^n_{*,m}\|^2.
	\end{align*}
\end{proof}
\subsection{Lemma 3}
For Algorithm~\ref{alg:vr_fed_rr} and Algorithm~\ref{alg:dvr_fed_rr} the following inequality holds:
\begin{align*}
\mathbb{E}\left[\| h_{t+1,m} - x^n_{*,m} \|^2 \right]\leq (1-\alpha)\mathbb{E}\| h_{t,m} - x^n_{*,m}\|^2+\alpha \mathbb{E}\| x^n_{t,m} - x^n_{*,m} \|^2.
\end{align*}
\begin{proof}
	Let us start from conditional expectation:
	\begin{align*}
	\mathbb{E}\left[\| h_{t+1,m} - x^n_{*,m} \|^2\mid x^n_{t,m},h_{t,m}\right] &= \mathbb{E}\left[\left\| h_{t,m} + \alpha q_{t,m}- x^n_{*,m} \right\|^2\mid x^n_{t,m},h_{t,m}\right]\\
	&\leq \| h_{t,m} - x_{*,m}^n \|^2 +2\alpha\left\langle h_{t,m} - x^n_{*,m},\mathbb{E}[q_{t,m}] \right\rangle+\alpha^2\mathbb{E}\left[ \|q_{t,m}\|^2 \right]\\
	&\leq \| h_{t,m} - x_{*,m}^n \|^2 +2\alpha\left\langle h_{t,m} - x^n_{*,m},x^n_{t,m} - h_{t,m} \right\rangle\\
	&+\alpha^2(\omega+1) \|x^n_{t,m} - h_{t,m}\|^2\\
	&\leq \| h_{t,m} - x_{*,m}^n \|^2 +2\alpha\left\langle h_{t,m} - x^n_{*,m},x^n_{t,m} - h_{t,m} \right\rangle\\
	&+\alpha\|x^n_{t,m} - h_{t,m}\|^2\\
	&=\| h_{t,m} - x_{*,m}^n \|^2 +\alpha\left\langle 2h_{t,m} - 2x^n_{*,m}+x^n_{t,m} - h_{t,m},x^n_{t,m} - h_{t,m} \right\rangle.
	\end{align*}
	Let us consider last term:
	\begin{align*}
	&\left\langle 2h_{t,m} - 2x^n_{*,m}+x^n_{t,m} - h_{t,m},x^n_{t,m} - h_{t,m} \right\rangle\\
	&= 	\left\langle h_{t,m} - x^n_{*,m}+  x^n_{t,m} - x^n_{*,m}, x^n_{t,m} - x^n_{*,m} - \left( h_{t,m} - x^n_{*,m} \right)  \right\rangle\\
	& = -\|h_{t,m} - x^n_{*,m}\|^2+\| x^n_{t,m} - x^n_{*,m}  \|^2.
	\end{align*}
\end{proof}
Using this result and previous inequlaity we get th following:
$$	\mathbb{E}\left[\| h_{t+1,m} - x^n_{*,m} \|^2\mid x^n_{t,m},h_{t,m}\right] \leq (1-\alpha)\| h_{t,m} - x^n_{*,m}\|^2+\alpha \| x^n_{t,m} - x^n_{*,m} \|^2.$$
Taking full expectation we finish the proof.
\subsection{Lemma 6}
For completeness we include the proof of important theorem introduced in~\citet{mishchenko2021proximal}. Suppose that each $f_i$ is $L$-smooth and $\mu$-strongly convex. Then the inner iterates satisfy 
$$\mathbb{E}\left[\left\|x_{t}^{i+1}-x_{*}^{i+1}\right\|^{2}\right] \leq(1-\gamma \mu) \mathbb{E}\left[\left\|x_{t}^{i}-x_{*}^{i}\right\|^{2}\right]-2 \gamma\left(1-\gamma L\right) \mathbb{E}\left[D_{f_{\pi_{i}}}\left(x_{t}^{i}, x_{*}\right)\right]+2 \gamma^{3} \sigma_{\mathrm{rad}}^{2}.$$
\begin{proof}
	By definition of $x^{i+1}_t$ and $x^{i+1}_*$, we have 
	\begin{align*}
	\mathbb{E}\left[\left\|x_{t}^{i+1}-x_{*}^{i+1}\right\|^{2}\right]=\mathbb{E} &\left[\left\|x_{t}^{i}-x_{*}^{i}\right\|^{2}\right]-2 \gamma \mathbb{E}\left[\left\langle\nabla f_{\pi_{i}}\left(x_{t}^{i}\right)-\nabla f_{\pi_{i}}\left(x_{*}\right), x_{t}^{i}-x_{*}^{i}\right\rangle\right] \\
	&+\gamma^{2} \mathbb{E}\left[\left\|\nabla f_{\pi_{i}}\left(x_{t}^{i}\right)-\nabla f_{\pi_{i}}\left(x_{*}\right)\right\|^{2}\right].
	\end{align*} 
	Note that the third term can be bounded as
	$$ \left\|\nabla f_{\pi_{i}}\left(x_{t}^{i}\right)-\nabla f_{\pi_{i}}\left(x_{t}^{i}\right)\right\|^{2} \leq 2 L \cdot D_{f_{\pi_{i}}}\left(x_{t}^{i}, x_{*}\right). $$
	Using the three-point identity we get 
	$$\left\langle\nabla f_{\pi_{i}}\left(x_{t}^{i}\right)-\nabla f_{\pi_{i}}\left(x_{*}\right), x_{t}^{i}-x_{*}^{i}\right\rangle=D_{f_{\pi_{i}}}\left(x_{*}^{i}, x_{t}^{i}\right)+D_{f_{\pi_{i}}}\left(x_{t}^{i}, x_{*}\right)-D_{f_{\pi_{i}}}\left(x_{*}^{i}, x_{*}\right).$$
	Combining these bounds we have 
	\begin{align*}
	\mathbb{E}\left[\left\|x_{t}^{i+1}-x_{*}^{i+1}\right\|^{2}\right] \leq \mathbb{E}\left[\left\|x_{t}^{i}-x_{*}^{i}\right\|^{2}\right] &-2 \gamma \cdot \mathbb{E}\left[D_{f_{\pi_{i}}}\left(x_{*}^{i}, x_{t}^{i}\right)\right]+2 \gamma \cdot \mathbb{E}\left[D_{f_{\pi_{i}}}\left(x_{*}^{i}, x_{*}\right)\right] \\
	&-2 \gamma\left(1-\gamma L\right) \mathbb{E}\left[D_{f_{\pi i}}\left(x_{t}^{i}, x_{*}\right)\right].
	\end{align*}
	Using $\mu$-strong convexity of $f_{\pi_i}$ , we derive
	$$\frac{\mu}{2}\left\|x_{t}^{i}-x_{*}^{i}\right\|^{2} \leq D_{f_{\pi_{i}}}\left(x_{*}^{i}, x_{t}^{i}\right).$$
	Using definition of shuffling radius we have 
	$$\mathbb{E}\left[D_{f_{\pi_{i}}}\left(x_{*}^{i}, x_{*}\right)\right] \leq \max _{i=1, \ldots, n-1} \mathbb{E}\left[D_{f_{\pi_{i}}}\left(x_{*}^{i}, x_{*}\right)\right]=\gamma^{2} \sigma_{\mathrm{rad}}^{2}.$$
	Putting all bounds together we get result. 
	
\end{proof}

\subsection{Proof of Theorem~\ref{thm:VR-CRR}}
\begin{proof}
	Let us define the Lyapunov function:
	$$\Psi_t = \|x_t-x_*\|^2+\frac{4 \eta^{2} \omega}{\alpha M}\frac{1}{M}\sum_{m=1}^{M}\left\| h_{t,m} - x^n_{*,m}  \right\|^2.$$
	Now we use Lemma 2 and Lemma 3:
	\begin{align*}
	\mathbb{E}\left[\Psi_{t+1}\right] &=  \mathbb{E}\|x_{t+1}-x_*\|^2+\frac{4 \eta^{2} \omega}{\alpha M}\frac{1}{M}\sum_{m=1}^{M}\mathbb{E}\left\| h_{t+1,m} - x^n_{*,m}  \right\|^2\\
	&\leq  \frac{2\eta^2}{M^2}\omega \sum_{m=1}^{M}\mathbb{E} \|x^n_{t,m} - x^n_{*,m}\|^2+\frac{2\eta^2}{M^2}\omega \sum_{m=1}^{M}\mathbb{E} \|h_{t,m} - x^n_{*,m}\|^2+(1-\eta)\mathbb{E}\|x_t-x_*\|^2\\
	&+\eta\frac{1}{M}\sum_{m=1}^{M}\mathbb{E}\|x^n_{t,m} - x^n_{*,m}\|^2+\frac{4 \eta^{2} \omega}{\alpha M}(1-\alpha)\sum_{m=1}^{M} \mathbb{E}\left\| h_{t,m} - x^n_{*,m}  \right\|^2+\frac{4 \eta^{2} \omega}{\alpha M}\alpha \sum_{m=1}^{M}\mathbb{E} \| x^n_{t,m} - x^n_{*,m} \|^2.
	\end{align*}
	Using Lemma 6 and Theorem 2 from~\citet{mishchenko2021proximal} we have 
	\begin{align*}
	\mathbb{E}\left[\Psi_{t+1}\right] &\leq \frac{4 \eta^{2} \omega}{\alpha M}\left(1-\frac{\alpha}{2}\right)\frac{1}{M}\sum_{m=1}^{M}\mathbb{E}\left\| h_{t,m} - x^n_{*,m}  \right\|^2+\left(1-\eta+\eta(1-\gamma\mu)^n+\frac{6\eta^2\omega}{M}(1-\gamma\mu)^n\right)\mathbb{E}\|x_t-x_*\|^2\\
	&+\left(\alpha+\eta+\frac{2\eta^2\omega}{M}\right)2\gamma^3\sigma^2_{rad}\left(\sum_{j=0}^{n-1}(1-\gamma\mu)^j\right)
	\end{align*}
	Using the condition $\eta \leq \min \left(1, \frac{\left(1-(1-\gamma \mu)^{n}\right) M}{12 \omega(1-\gamma \mu)^{n}}\right)$ we have
	\begin{align*}
	\mathbb{E}\left[\Psi_{t+1}\right] \leq \max \left(1-\frac{\alpha}{2},1-\frac{\eta\left(1-(1-\gamma\mu)^n\right)}{2}\right)  \mathbb{E}\left[\Psi_t\right]+\left(\alpha+\eta+\frac{2\eta^2\omega}{M}\right)2\gamma^3\sigma^2_{rad}\left(\sum_{j=0}^{n-1}(1-\gamma\mu)^j\right)
	\end{align*}
	Note that we have the following inequality:
	\begin{align*}
	1-\gamma\mu \leq 1 - \frac{1}{2}\eta \left(1-(1-\gamma\mu)^n\right)\\
	-\gamma\mu \leq -\frac{1}{2}\eta \left(1-(1-\gamma\mu)^n\right)\\
	\gamma\mu \geq \frac{1}{2}\eta \left(1-(1-\gamma\mu)^n\right).
	\end{align*}
	We have it since $0<\eta\leq 1$ and $n>1$, so we have 
	\begin{align*}
	\gamma\mu &\geq \frac{1}{2} \left(1-(1-\gamma\mu)^n\right)\\
	\gamma\mu &\geq \frac{1}{2} \left(1-(1-\gamma\mu)\right)\\
	1&\geq \frac{1}{2}.
	\end{align*}
	Unrolling this recursion finishes the proof.
\end{proof}
\subsection*{Lemma 4}
For completeness we include the proof of important theorem introduced in~\citet{malinovsky2021random}.
Suppose that the functions $f_1, \ldots, f_n$ are $\mu$-strongly convex and $L$-smooth. Fix constant $0 < \delta < 1$. If the
stepsize satisfies $\gamma \leq \frac{\delta}{L} \sqrt{\frac{\mu}{2 n L}}$ and if number of functions is sufficiently big:
$$n>\log \left(\frac{1}{1-\delta^{2}}\right) \cdot\left(\log \left(\frac{1}{1-\gamma \mu}\right)\right)^{-1}$$
and
$$\delta^{2} \leq(1-\gamma \mu)^{\frac{n}{2}}\left(1-(1-\gamma \mu)^{\frac{n}{2}}\right) .$$
$$\mathbb{E}\left[\left\|x^n_{t}-x^n_{*}\right\|^{2}\right] \leq (1-\gamma \mu)^{\frac{n}{2}} \mathbb{E}\left[\left\|x_{t}-x_{*}\right\|^{2}\right] ,$$
then we have

$$\mathbb{E}\left[\left\|x^n_{t}-x^n_{*}\right\|^{2} \mid x_{t}\right] \leq(1-\gamma \mu)^{n}\left\|x_{t}-x_{*}\right\|^{2}+\frac{\gamma^{3} L n}{2} \sigma_{*}^{2}\left(\sum_{i=0}^{n-1}(1-\gamma \mu)^{i}\right)$$
\begin{proof}
	We start from Theorem 1 in~\citet{mishchenko2020random}:
	$$\mathbb{E}\left[\left\|x^n_{t}-x^n_{*}\right\|^{2} \mid x_{t}\right] \leq(1-\gamma \mu)^{n}\left\|x_{t}-x_{*}\right\|^{2}+\frac{\gamma^{3} L n}{2} \sigma_{*}^{2}\left(\sum_{i=0}^{n-1}(1-\gamma \mu)^{i}\right)$$
	Using property of geometric progression we can have an upper bound $\sum_{i=0}^{n-1}(1-\gamma \mu)^{i} \leq \frac{1}{\gamma \mu}:$
	\begin{align*}
	\mathbb{E}\left[\left\|x^n_{t}-x^n_{*}\right\|^{2} \mid x_{t}\right] &\leq(1-\gamma \mu)^{n}\left\|x_{t}-x_{*}\right\|^{2}+\frac{\gamma^{2} L n}{2 \mu} \sigma_{*}^{2}.
	\end{align*}
	Using Lemma 1 in~\citet{malinovsky2021random} we get
	\begin{align*}
	\mathbb{E}\left[\left\|x^n_{t}-x^n_{*}\right\|^{2} \mid x_{t}\right] &\leq\left((1-\gamma \mu)^{n}+\frac{2 \gamma^{2} L^{3} n}{\mu}\right)\left\|x_{t}-x_{*}\right\|^{2}.
	\end{align*}
	Let us use $\gamma \leq \frac{\delta}{L} \sqrt{\frac{\mu}{2 n L}}$. To get convergence we need 
	$$(1-\gamma \mu)^{n}+\delta^{2}<1.$$ 
	This leads to the following inequality: 
	$$n>\log \left(\frac{1}{1-\delta^{2}}\right) \cdot\left(\log \left(\frac{1}{1-\gamma \mu}\right)\right)^{-1}.$$
	Also assume 
	$$\delta^{2} \leq(1-\gamma \mu)^{\frac{n}{2}}\left(1-(1-\gamma \mu)^{\frac{n}{2}}\right) .$$
	Putting this into bound finishes the proof.
\end{proof}
\subsection{Lemma 5}
For completeness we include the proof of important lemma introduced in~\citet{malinovsky2021random}.

Assume that each $f_i$ is $L$-smooth and convex. If we apply the linear perturbation reformulation~\ref{perty}, then the variance of reformulated problem satisfies the following inequality:
$$ \tilde{\sigma}_{*}^{2} \leq 4L^2\|y_t-x_*\|^2.$$
\begin{proof}
	$$\tilde{\sigma}_{*}^{2} =\frac{1}{n} \sum_{i=1}^{n}\left\|\nabla f_{i}\left(x_{*}\right)-\nabla f_{i}\left(y_{t}\right)+\nabla f\left(y_{t}\right)-\nabla f\left(x_{*}\right)\right\|^{2}$$	
	Using Young's inequality we have 
	\begin{align*}
	\tilde{\sigma}_{*}^{2} &\leq \frac{1}{n} \sum_{i=1}^{n}\left(2\left\|\nabla f_{i}\left(y_{t}\right)-\nabla f_{i}\left(x_{*}\right)\right\|^{2}+2\left\|\nabla f\left(y_{t}\right)-\nabla f\left(x_{*}\right)\right\|^{2}\right)\\
	& \leq \frac{1}{n} \sum_{i=1}^{n} 4 L_{i} D_{f_{i}}\left(y_{t}, x_{*}\right)+\frac{1}{n} \sum_{i=1}^{n} 4 L D_{f}\left(y_{t}, x_{*}\right) \\
	& \leq 4 L D_{f}\left(y_{t}, x_{*}\right)+4 L D_{f}\left(y_{t}, x_{*}\right) \\
	&=8 L D_{f}\left(y_{t}, x_{*}\right) \\
	& \leq 4 L^{2}\left\|y_{t}-x_{*}\right\|^{2}
	\end{align*}
\end{proof}

\subsection{Proof of Theorem~\ref{thm:double}}
The proof is similar to the proof of~\ref{thm:VR-CRR}.
\begin{proof}
	Let us define the Lyapunov function:
	$$\Psi_t = \|x_t-x_*\|^2+\frac{4 \eta^{2} \omega}{\alpha M}\frac{1}{M}\sum_{m=1}^{M}\left\| h_{t,m} - x^n_{*,m}  \right\|^2.$$
	Now we use Lemma 2 and Lemma 3:
	\begin{align*}
	\mathbb{E}\left[\Psi_{t+1}\right] &=  \mathbb{E}\|x_{t+1}-x_*\|^2+\frac{4 \eta^{2} \omega}{\alpha M}\frac{1}{M}\sum_{m=1}^{M}\mathbb{E}\left\| h_{t+1,m} - x^n_{*,m}  \right\|^2\\
	&\leq  \frac{2\eta^2}{M^2}\omega \sum_{m=1}^{M}\mathbb{E} \|x^n_{t,m} - x^n_{*,m}\|^2+\frac{2\eta^2}{M^2}\omega \sum_{m=1}^{M}\mathbb{E} \|h_{t,m} - x^n_{*,m}\|^2+(1-\eta)\mathbb{E}\|x_t-x_*\|^2\\
	&+\eta\frac{1}{M}\sum_{m=1}^{M}\mathbb{E}\|x^n_{t,m} - x^n_{*,m}\|^2+\frac{4 \eta^{2} \omega}{\alpha M}(1-\alpha)\sum_{m=1}^{M} \mathbb{E}\left\| h_{t,m} - x^n_{*,m}  \right\|^2+\frac{4 \eta^{2} \omega}{\alpha M}\alpha \sum_{m=1}^{M}\mathbb{E} \| x^n_{t,m} - x^n_{*,m} \|^2.
	\end{align*}
	Using Lemma 5 and Theorem 3 from~\citet{malinovsky2021random} we have 
	\begin{align*}
	\mathbb{E}\left[\Psi_{t+1}\right] &\leq \frac{4 \eta^{2} \omega}{\alpha M}\left(1-\frac{\alpha}{2}\right)\frac{1}{M}\sum_{m=1}^{M}\mathbb{E}\left\| h_{t,m} - x^n_{*,m}  \right\|^2+\left(1-\eta+\eta(1-\gamma\mu)^{\frac{n}{2}}+\frac{6\eta^2\omega}{M}(1-\gamma\mu)^{\frac{n}{2}}\right)\mathbb{E}\|x_t-x_*\|^2\\
	&+\left(\alpha+\eta+\frac{2\eta^2\omega}{M}\right)2\gamma^3L \sum_{m=1}^{M}\left\|\nabla F_{m}\left(x_{*}\right)\right\|^{2}\left(\sum_{j=0}^{n-1}(1-\gamma\mu)^j\right)
	\end{align*}
	Using the condition $\eta \leq \min \left(1, \frac{\left(1-(1-\gamma \mu)^{\frac{n}{2}}\right) M}{12 \omega(1-\gamma \mu)^{\frac{n}{2}}}\right)$ we have
	\begin{align*}
	\mathbb{E}\left[\Psi_{t+1}\right] &\leq \max \left(1-\frac{\alpha}{2},1-\frac{\eta\left(1-(1-\gamma\mu)^{\frac{n}{2}}\right)}{2}\right)  \mathbb{E}\left[\Psi_t\right]\\
	&+\left(\alpha+\eta+\frac{2\eta^2\omega}{M}\right)2\gamma^3L \sum_{m=1}^{M}\left\|\nabla F_{m}\left(x_{*}\right)\right\|^{2}\left(\sum_{j=0}^{n-1}(1-\gamma\mu)^j\right)
	\end{align*}
	Unrolling this recursion as we did previously finishes the proof.
\end{proof}

\end{document}